%% file: main.tex
\documentclass[11pt,a4paper,final]{article}
\usepackage{newunicodechar}
\newunicodechar{−}{\ensuremath{-}}
\usepackage[left=1in,right=1in,top=1in,bottom=1in]{geometry}
\usepackage{palatino} 
\usepackage{authblk}
\PassOptionsToPackage{sort&compress}{natbib}
\usepackage[]{natbib}

\usepackage{booktabs} 
\usepackage[ruled]{algorithm2e} 
\usepackage{tikz}
\usetikzlibrary{arrows}
\usetikzlibrary{shapes.multipart}
\usepackage{longtable}
\usepackage{caption}

\SetAlFnt{\small}
\SetAlCapFnt{\small}
\SetAlCapNameFnt{\small}
\SetAlCapHSkip{0pt}
\IncMargin{-\parindent}
\usepackage[all]{nowidow}

\usepackage[font=small,labelfont=bf]{caption}
\usepackage[utf8]{inputenc} 
\usepackage[T1]{fontenc}    
\usepackage{hyperref}       
\hypersetup{
    colorlinks=true,
    citecolor=purple,
    linkcolor=blue,
    filecolor=magenta,
    urlcolor=cyan,
    pdfpagemode=FullScreen,
    }
\usepackage{url}            
\usepackage{booktabs}       
\usepackage{amsfonts}       
\usepackage{nicefrac}       
\usepackage{microtype}      
\usepackage{xcolor}         
\usepackage{amsmath}
\usepackage{mathtools}
\usepackage{amsthm}
\usepackage{amsmath}
\usepackage{amssymb}
\usepackage{bbm}
\usepackage{tikz} 
\usepackage{wrapfig}
\usepackage{floatrow}
\usepackage{enumitem}

\usepackage[suppress]{color-edits}
\usepackage{caption}
\usepackage{subcaption}

\usepackage{xcolor}

\title{Tradeoffs between objective outcomes and human enjoyment of work in human-AI task allocation}

\begin{document} 
\author[1,2]{Katherine M. Collins\footnote{Corresponding author: {katiemc@mit.edu}}}
\author[2]{Elizabeth Mieczkowski}
\author[3]{Raja Marjieh}
\author[1]{\\Marlene Berke}
\author[1]{Joshua B. Tenenbaum}
\author[4]{Max Kleiman-Weiner}
\author[$\dagger$\,5]{\\Ilia Sucholutsky}
\author[$\dagger$\,2]{Thomas L. Griffiths}
\affil[1]{Massachusetts Institute of Technology}
\affil[2]{Princeton University}
\affil[3]{Harvard University}
\affil[4]{University of Washington}
\affil[5]{Purdue University}
\affil[$\dagger$]{Co-senior authors.}

\maketitle

\begin{abstract}

For people, tasks at work are defined by more than their outcome: they are colored by the experience of engaging in that task. As AI systems become increasingly capable of performing human-level work they may be deployed to tasks that people find enjoyable while simultaneously creating comparatively more aversive tasks such as monitoring AI outputs. This creates a thorny tradeoff for managers: whether work should be assigned purely based on the expected performance of the workers in question, or also account for the subjective experience of the person potentially doing the task. Most economic accounts focus on expected utility calculations based on the former. Here, we assess how people balance objective outcomes and more subjective qualities like enjoyment. In a series of large-scale ($N=800$) behavioral studies in which people allocate tasks to either an AI or human worker we find that people's allocation decisions are best captured by a model that systematically balances relative objective performance with the human worker's expected enjoyment, weighting human experience higher when a task is framed as inducing less enjoyment and adjusting based on the stakes of the task. Given rising consideration of large language models (LLMs) for automated managerial decision making, we compare a set of LLMs to people's choices. We find that many models also deviate from purely objective decisions, particularly when the project stakes are low, but there is wide heterogeneity across models as to how much and in what manner they weigh enjoyment. These results raise questions about what role human experience may play in decisions around where and how to integrate AI into workplaces. Integration may be jagged across settings due to consideration not only of the relative capabilities of workers, but also the human experience of work, the valence of that experience (e.g., more strongly weighing aversion), the context (e.g., stakes), and who is making or guiding such choices (human or LLM).

\end{abstract}

\newpage 

\noindent \textit{``There is a real joy in doing mathematics.''}
\begin{flushright}-- William P. Thurston, {\em On Proof and Progress in Mathematics} \cite{thurston1995proof}
\end{flushright}\vspace{2mm}

\noindent \textit{``Actually, the laboring man has not leisure for true integrity day by day [...] his labor would be depreciated in the market. He has no time to be anything but a machine.''}
\begin{flushright}-- Henry David Thoreau, {\em Walden}~\cite{thoreau1854walden}
\end{flushright}\vspace{2mm}

\begin{figure}[h!]
    \centering
    \includegraphics[width=0.9\linewidth]{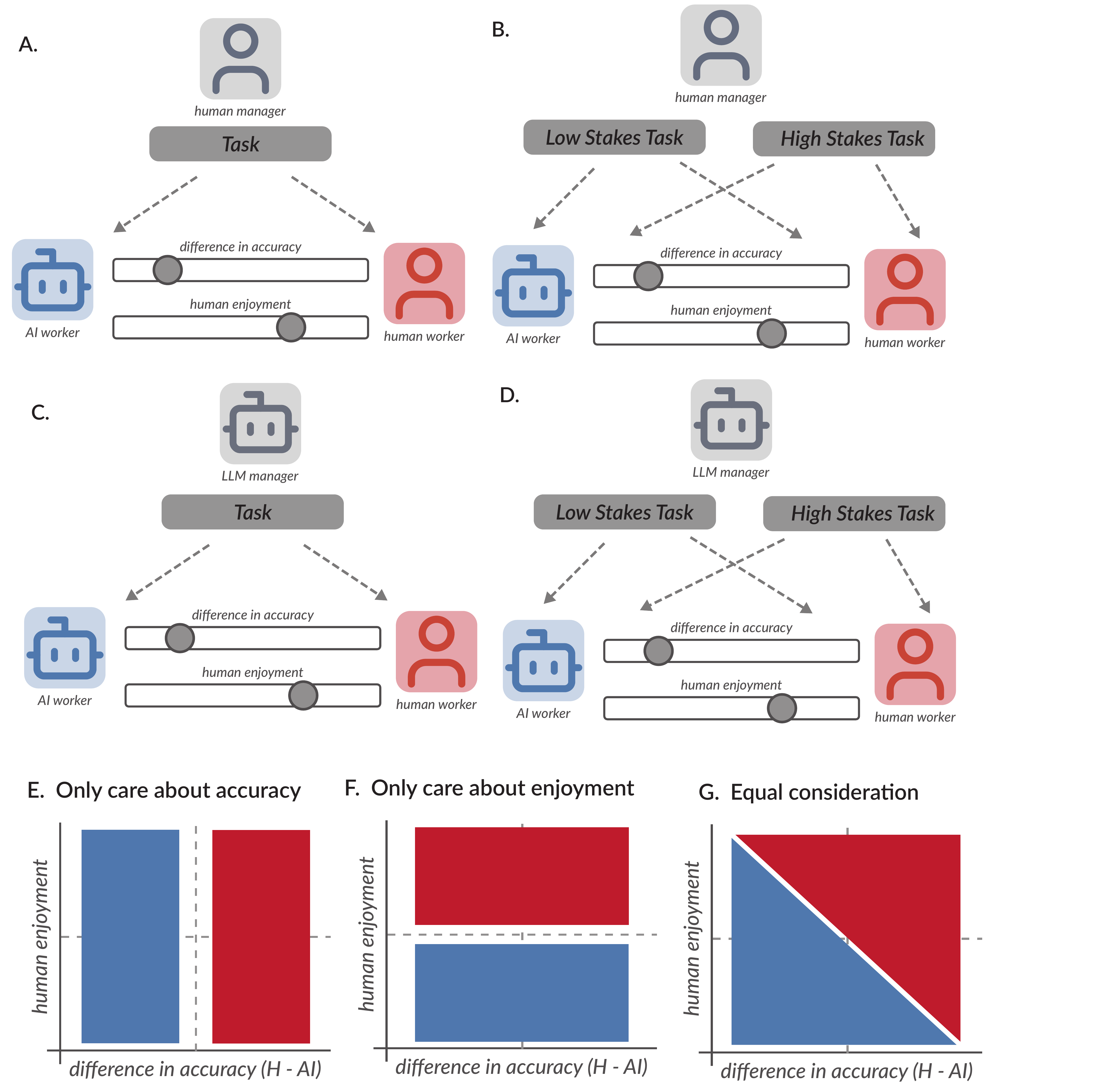}
    \caption{\textbf{Tradeoffs in task allocation.} \textbf{(A)-(B)} In our setup, a manager needs to decide whether to assign a task to either an AI worker or human. The manager may know the relative performance of the two workers and the expected enjoyment of the human worker. When these conflict (e.g., the AI worker is better at the task but the human enjoys it), the manager needs to make a tradeoff. In increasingly hybrid workplaces, not only are workers AI but the managers too (\textbf{(C)-(D)}). In Experiment 1, both people \textbf{(A)} and LLMs \textbf{(C)} make decisions about which worker to assign a task to. In Experiment 2, human\textbf{(B)} and LLM managers \textbf{(D)} are also told the stakes of the task being assigned. Managers may follow different allocation policies, for instance, weighing only accuracy \textbf{(E)}, enjoyment \textbf{(F)}, or an equal combination \textbf{(G)}. Red means a task was assigned to a human worker; blue means assigned to an AI worker.}
    \label{fig:overview-schematic}
\end{figure}

The advent of new technologies often brings with it questions of impact on human labor~\citep{smith1776wealth, acemoglu2026building, cockburn2018impact, eloundou2024gpts, bresnahan1995general, dell2023navigating, solow1957, brynjolfsson2025generative, milewski1997delegating}. From mechanized looms to ATMs, debates arise as to whether and how to integrate such technology into human workplaces. This discussion has reached a fever pitch in the discourse around artificial intelligence (AI), with growing concerns for the impact of AI not only on job displacement but on human \textit{experience} of work~\citep{ranjit2026we, collins2026humanagencyinitialhumanai, shao2025future, sadeghian2024soul}. AI capabilities continue to extend to an ever-widening range of tasks, some of which people may find inherently enjoyable~\citep{ranjit2026we}. Accordingly, AI could automate many of the tasks that people enjoy the most, while pushing them into roles, such as verifying outputs rather than creating them, which people may find more aversive~\citep{simkute2025ironies}. Integrating AI into human work thus raises the potentially thorny tradeoff of balancing productivity with the fundamentally human \textit{experience} of the work (Figure~\ref{fig:overview-schematic}A-D). 

This tradeoff has been called out before. In 1973, when discussing the integration of information technology into work, Herbert Simon asked whether ``there should be a deliberate sacrifice of effectiveness and efficiency in order to make work itself a rewarding and enjoyable part of human life''~\citep{simon1973applying}. Under classic decision-theoretic models~\citep{von1947theory}, managers whose utility depends solely on task outcomes should allocate work to maximize expected effectiveness or efficiency (Figure~\ref{fig:overview-schematic}E). Prior work has examined how people allocate tasks based on collaborators' capabilities and expected performance \citep{mieczkowski2026people, marjieh2024task, xiang2023collaborative}. Yet workers and managers alike also consider other experiential factors when allocating and selecting work, e.g., \citep{cai2023incentive, stern2004scientists, bandiera2009social, mas2017valuing, cassar2018nonmonetary}. Systematically characterizing and formalizing such tradeoffs remains comparatively underexplored. Such efforts are important if we want to anticipate the kinds of choices people may make (and reactions they may have to others' choices) when deciding how to allocate tasks between people and AI systems in increasingly hybrid workplaces. 

In current and future workplaces, the navigator of such tradeoffs may not always be human. AI systems, particularly large language models (LLMs), may increasingly be consulted as ``thought partners''~\citep{collins2024building} by people making allocation decisions, used to simulate how people might make choices in economic settings~\citep{horton2023large, manning2026general, liu2024large, gao2024large, sun2026llms}, or even take on the role of task allocator~\citep{li2026large, prunty2026reverse, tomavsev2026intelligent,guggenberger2023task, armstrong2024silicon} (Figure ~\ref{fig:overview-schematic}C-D). Yet, such systems do not have any direct experience with what it means to have (or lack) enjoyment at work. As LLMs operate under fundamentally different training regimes and experiences to people~\citep{mccoy2024embers} that could percolate into such decisions, it is therefore paramount that we also measure and model the kinds of profiles LLMs may employ in making allocation tasks (e.g., Figure~\ref{fig:overview-schematic}E-G) and where and how those differ from people's intuitions.

To explore how people and LLMs manage tradeoffs between objective outcomes and subjective experience, we designed a simple parameteric allocation task. Decision makers (human participants and LLMs) were tasked with playing the role of a manager and presented with a series of ``special projects.'' For each project, they are told the expected accuracy of an AI system compared to a human worker and the expected enjoyment of that human worker relative to a task they otherwise could be doing. We first considered the case where decisions are made for tasks with unspecified stakes (Figure~\ref{fig:overview-schematic}A,C). We then assessed how stakes influence decisions (Figure~\ref{fig:overview-schematic}B,D). We ran both pre-registered experiments with human participants recruited from a crowdsourcing platform (Prolific~\citep{palan2018prolific}; see Methods) and repeated the decisions across a series of $21$ LLMs across $7$ model families making the allocation decisions using a lightly modified version of the same instructions given to people (see S.I.~\ref{sec:si-llm-prompts}).

\section*{Results}

\paragraph{People and many LLMs are willing to sacrifice objective performance for subjective preferences when assigning tasks between human and AI workers.} 

\begin{figure}[h!]
    \centering
    \includegraphics[width=1.0\linewidth]{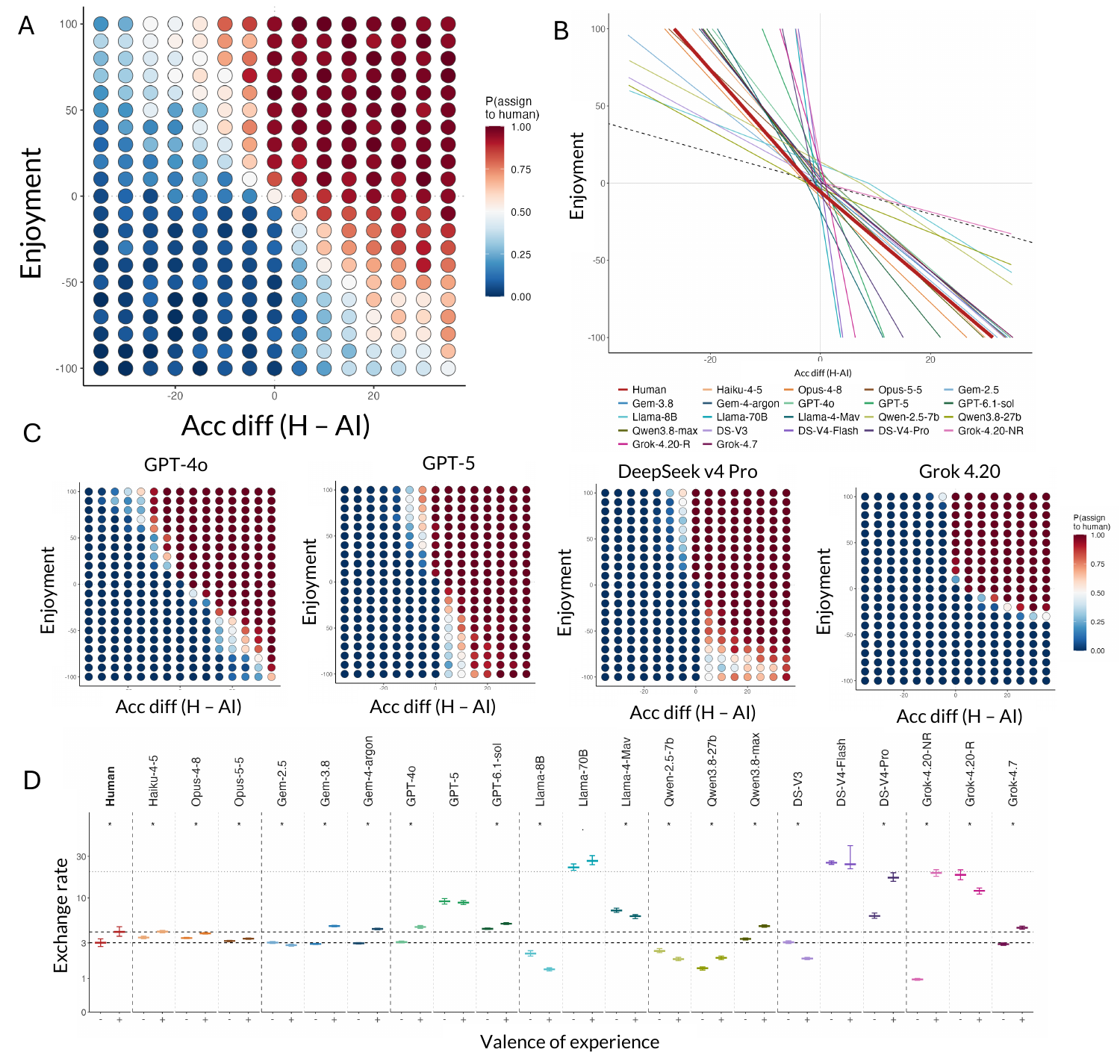}
    \caption{\textbf{Tradeoffs of objective outcomes and subjective experience of work for humans and LLMs}. \textbf{(A)} Assignments to human worker or AI, averaged over the $40$ human participants who responded for each trial. Darker red indicates a higher proportion of tasks assigned to the human worker, while darker blue indicates a lower proportion assigned to the human worker and thus a higher proportion assigned to AIs.  \textbf{(B)} $50-50$ contour line for piecewise logistic models fit to people's choices (red) versus fit separately to the choices of different LLMs. \textbf{(C)} Example allocation profiles of LLMs,  highlighting the diversity in extent and manner of weighing human experience relative to objective outcomes. For instance, GPT-5 typically assigns the task to the more accurate worker, but slightly weighs human experience in decisions; in contrast, Grok 4.20 primarily assigns to tasks to people when the human worker is both better at the task and enjoys the work, with some systematic preference in favoring human experience only when the person actively does not enjoy the task. Full set included in S.I.~\ref{sec:si-llm-full}. \textbf{(D)} Average exchange rates for the $\rho_h < 0$ and $\rho_h > 0$ regions fit in a piecewise logistic model to sets of choices from each agent (human or LLM). Error bars depict bootstrapped 95\% CIs. Higher exchange rate means human enjoyment is weighed less than accuracy; an exchange rate of one indicates that human experience and expected accuracy are weighed equally. The dotted line at $20$ indicates the highest ER that the design can distinguish, since trial accuracy differs in steps of $5$ and the maximum enjoyment is $100$; values with ER $> 20$ only ever assign on accuracy unless the workers have equivalent expected accuracy (see Methods). Stars indicate significance, with $^{***} q < 0.001$, $^{**} q < 0.01$, $^{*} q< 0.05$ (reported as $q$ based on Benjamini-Hochberg multiple hypothesis correction).}
    \label{fig:tradeoff-main-no-stakes}
\end{figure}

We recruited $200$ participants from Prolific in a pre-registered study to decide whether to allocate tasks to a person or AI system, after having been provided both agents' expected accuracy and the human worker's expected relative enjoyment (see Methods and example interface in S.I. Figure~\ref{fig:experiment-interface}A). We model people's choices with a logistic regression that accounts for both  human enjoyment ($\rho_{h}$) and the difference in agent capabilities ($x_h - x_{AI}$, where $x_h$ and $x_{AI}$ are the human and model expected accuracy on the task respectively) when allocating tasks between people and AI: $\Pr(\text{assign human}) = \sigma(\beta_\text{offset} + \beta_{\text{acc}} \times (x_h - x_{AI}) + \beta_{\text{enj}} \times \rho_{h})$. People assigned tasks in a way that is consistent with a model that weighs both agents' relative accuracy and human enjoyment in their allocation decision (see Figure~\ref{fig:tradeoff-main-no-stakes}A; both predictors were reliably positive: $\beta_{\text{acc}} = 0.09$, SE $= 0.006$, $t = 14.4$, $p < 0.001$; $\beta_{\text{enj}} = 0.03$, SE $= 0.002$, $t = 14.5$, $p < 0.001$). Participant-cluster-robust Wald tests (see Methods) confirmed that both features contribute, either adding accuracy difference to a model that only has enjoyment ($F(1, 199) = 207.7$, $p < 0.001$) or adding enjoyment to a model that only has accuracy difference ($F(1, 199) = 209.9$, $p < 0.001$).

However, the tradeoff is nuanced and asymmetric. We next estimated the ``exchange rate'' (ER) implied by people's choices, that is, how many effective points of enjoyment are needed to offset a one point difference in accuracy to shift their allocation (computed as $\beta_{acc} / \beta_{enj}$). 
The average number of enjoyment points needed to offset a one point difference in accuracy is $3.46$ ($95\%$ CI [$3.11, 3.87$]; model fits were bootstrapped over participant judgments, see Methods). This means that approximately $35$ points of enjoyment are needed to offset a $10$ point swing in relative accuracy between the human worker and AI to shift someone's allocation. In a secondary analysis, we additionally assessed whether people's tradeoffs were sensitive to the valence of subjective experience. Fitting a piecewise logistic regression that changes slope at $p_h = 0$ revealed that people weigh subjective experience more when the subjective experience is framed negatively (ER $= 3.01$, $95\%$ CI [$2.69, 3.41$] for $\rho_\text{h} < 0$) compared to positively (ER $= 4.05$, $95\%$  CI [$3.53, 4.63$] for $\rho_\text{h} > 0$; see S.I. Figure~\ref{fig:contour-plots-no-stakes}). This has parallels to other forms of loss aversion~\citep{prospectTheory1979, peterson2021using}, but here, potentially operates over a managers' decision about someone else. 

Testing LLMs on the same decision tasks revealed that most models also trade off the expected human experience of work with the relative accuracy of workers (Figure~\ref{fig:tradeoff-main-no-stakes}B). This is a notable finding for organizations that may consider using LLMs as part of task allocation: models do not always allocate tasks based on purely the objective expected value when allocating work. However, there is wide heterogeneity across models, even within model families, in whether and how much they weigh human enjoyment (Figure~\ref{fig:tradeoff-main-no-stakes}C). $13$ of the $21$ models make choices consistent with an exchange rate significantly different from people; $8$ of the $13$ have a higher ER than people, needing more enjoyment points to offset a difference in accuracy (these LLMs' average ER range from approximately $4$ to $25$ enjoyment points relative to accuracy, compared to an average of $3.5$ for people; see Figure~\ref{fig:tradeoff-main-no-stakes}D). Additionally, while $18$ of the $21$ models show a significant ($q < 0.05$, based on a Benjamini-Hochberg correction; see Methods) asymmetry in how they weigh the valence of experience, $6$ of the $18$ show the opposite asymmetry to people, weighing positive valenced experience comparatively stronger than negative valenced (see S.I.~\ref{sec:si-llm-full} for all LLM choices). 

\paragraph{People and LLMs' tradeoffs are sensitive to project stakes.} 

\begin{figure}[h!]
    \centering
    \includegraphics[width=1.0\linewidth]{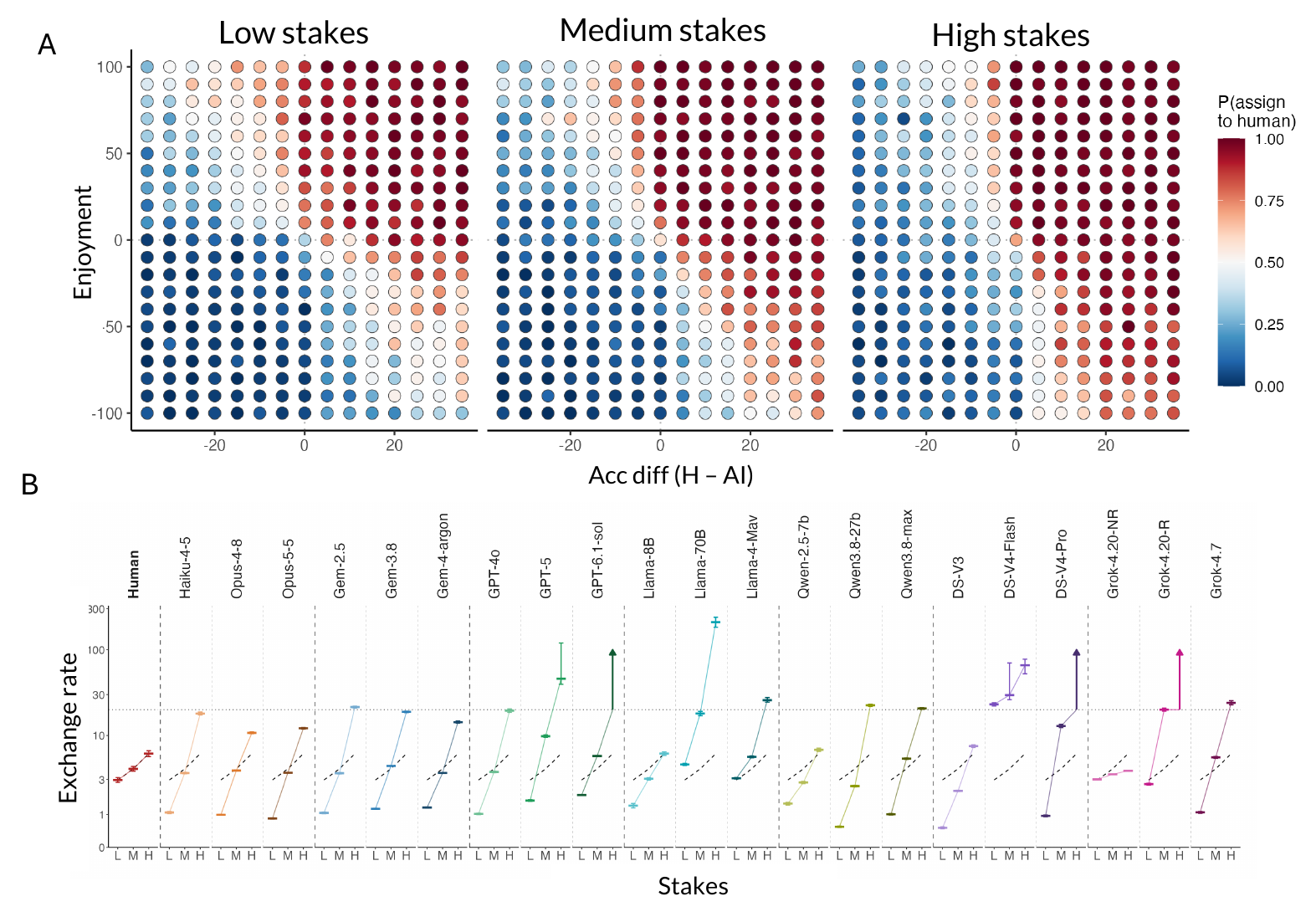}
        \caption{\textbf{Tradeoffs of objective outcomes and subjective experience of work as a function of project stakes for humans and LLMs}. \textbf{(A)} Assignments to human worker or AI, averaged over the $40$ human participants who responded for each trial, decomposed by whether the task was framed as low (left), medium (middle), or high (right) stakes.  
        \textbf{(B)} Relative exchange rates (ERs; ratio of the coefficient on enjoyment $\beta_{acc} / \beta_{enj}$) for each stakes setting, between people and different LLMs. Higher exchange rates means that relative accuracy is weighed even more strongly than enjoyment. LLM ERs are depicted adjacent to dashed red lines indicating the human ERs. Error bars depict bootstraps over $10,000$ model fits to resampled allocations (sampled over participants for humans and over rollouts for LLMs).  Note three LLMs (GPT-6.1-Sol, DeepSeek-V4-Pro, and Grok 4.20 (reasoning)) had unbounded ERs as they always assigned on accuracy and had little rollout variability (see Methods). } 
    \label{fig:tradeoff-main-stakes}
\end{figure}

In practice, the tasks that a manager needs to allocate might differ in their importance to the organization and individuals that might be impacted by the work of a project (Figure~\ref{fig:overview-schematic}B). If a project is extremely high stakes, one may be much more focused on making decisions that favor yielding the best outcomes on the project (which might involve overlooking the enjoyment of the worker involved). In contrast, for a project that is low stakes, one might be more open to weighing the experience of doing the work (if the outcome matters less). 

We recruited a new set of $600$ participants in a second pre-registered study to engage in the same kind of allocation decisions, where now each special project was also tagged as being low, medium, or high stakes (see choices in Figure~\ref{fig:tradeoff-main-stakes}A; further details in Methods and example interfaces in S.I. Figure~\ref{fig:experiment-interface}B-C). We extended our computational model to account for stakes (see Methods). Allowing both relative accuracy ($\beta_{\text{acc}}$) and enjoyment ($\beta_{\text{enj}}$) weights to vary by stakes level improved on a model holding
them fixed ($F(4, 599) = 90.0$, $p < 0.001$), and the way accuracy and enjoyment combined also shifted across levels (three-way stakes $\times$ accuracy $\times$ enjoyment; $\chi^2(2) = 15.83$, $p < 0.001$).
While enjoyment mattered at each stakes level ($\beta_{\text{low}} = 0.027$, $\beta_{\text{med}} = 0.026$,
$\beta_{\text{high}} = 0.018$; all $p < 0.001$), its relative weighting mattered more for lower than higher stakes. Both accuracy interactions were positive, from low to medium ($t=9.0$, $p < 0.001$) and low to high ($t=8.3$, $p < 0.001$). While the weight on enjoyment significantly decreased from low to high stakes  ($t=-11.4$, $p < 0.001$), the weight on enjoyment from low to medium stakes moved in the predicted direction, it was not significant ($t=-1.2$, $p=0.22$; see Methods). This suggests the rise in ER from low to medium stakes stems from higher weight on accuracy. Computing the exchange rate per stake setting revealed a mononotic ordering over relative enjoyment versus accuracy weighting (Figure~\ref{fig:tradeoff-main-stakes}B). That is, people's choices were best captured by a model with an ER of $2.96$ enjoyment points per accuracy point at low stakes ($95\%$ CI $[2.77, 3.16]$) to $4.03$ at medium ($95\%$ CI $[3.78, 4.30]$) to $6.11$ at high ($95\%$ CI $[5.66, 6.61]$). Additionally, a secondary analysis into asymmetry in valence revealed people favored subjective experience for human workers who would face \textit{less} enjoyment by doing the task, when the stakes were medium or high ($1.03$ with $95\%$ CI $[0.47, 1.60], p < 0.001$ and $2.37$ with $95\%$ CI $[1.01, 3.86], p < 0.001$ for the difference in ER between the $\rho >0$ and $\rho < 0$ regions for the medium and high stakes settings, respectively), but no significant difference when the stakes were low ($0.28$ with $95\%$ CI $[−0.10, 0.68]$; see S.I. Figure~\ref{fig:exchange-rate-asymm-stakes}).  

All LLMs exhibit a monotonically stronger weighting of accuracy over enjoyment as stakes rise (see Figure~\ref{fig:tradeoff-main-stakes}B). However, the dynamic range over which enjoyment and accuracy were weighed was significantly wider for most LLMs than people. $20$ of the $21$ LLMs had a higher dynamic range than people ($q < 0.05$). The median difference in average low stakes ER versus high stakes ER was $3.2$ for people compared to a median of $18.1$ for the LLMs (note: three LLMs of the $20$ LLMs, GPT-6.1-Sol, DeepSeek-v4-Pro, and Grok 4.20 (reasoning), always allocated by accuracy for the high stakes setting and have unbounded ER and accordingly an even higher effective dynamic range). $17$ of the $21$ LLMs weigh enjoyment over accuracy for low stakes more strongly than people ($q < 0.05$ on ER), and conversely, $19$ of $21$ models weigh accuracy over enjoyment more strong than people for high stakes ($q < 0.05$ on ER). Moreover, at high stakes, $11$ of $21$ LLMs choices imply an ER above $20$, suggesting that the logistic model fit favors the more accurate worker in any possible conflict with accuracy and enjoyment (see Figure~\ref{fig:tradeoff-main-stakes}B); $5$ models never allocated tasks in a way that overrode objective outcomes based on enjoyment
(see Methods). We include full scatter and contour plots on fit logistic models in S.I. ~\ref{sec:si-llm-full}.

\begin{figure}[h!]
    \centering
    \includegraphics[width=\linewidth]{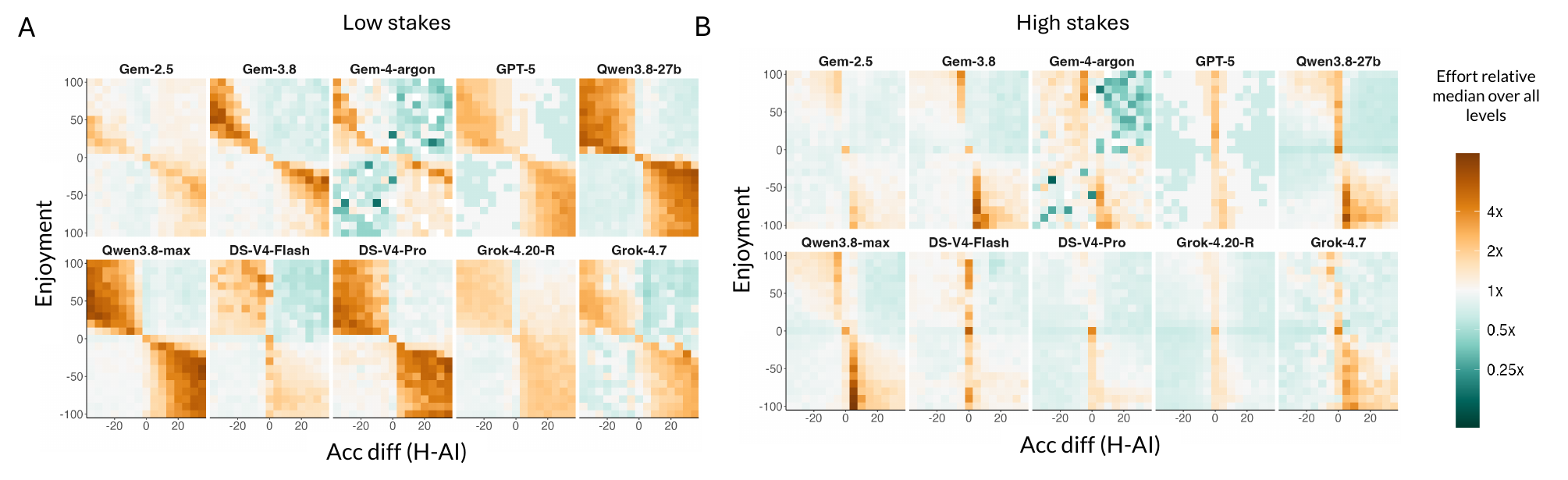}
    \caption{\textbf{Reasoning token consumption across allocation decisions}. Relative token consumption by different LLMs across the trials, for the low stakes (\textbf{(A)}) and high stakes (\textbf{(B)}) trials. Token consumptions is normalized within each LLM and across all allocation decisions. Darker orange means that more tokens were consumed than the median token consumption over all trials for that model; darker green means relatively less tokens were consumed by that model compared to the median over all of its trials. The full set of experimental conditions (including medium stakes and the unspecified stakes condition) included in S.I. Figure~\ref{fig:reasoning-tokens-full}.}
    \label{fig:reasoning-tokens}
\end{figure}

\paragraph{Navigating tradeoffs comes with hidden costs.} 

How such tradeoffs are navigated may have an impact not only on those downstream of the choice (e.g., human worker) but on the navigator. In exploratory analyses, we analyzed the number of reasoning tokens (an approximation of model effort~\citep{de2025cost}) that LLMs consumed when undertaking each decision problem. For the subset of LLMs tested that engage in reasoning and permit access to the number of reasoning tokens used, we observed that many of these models consumed more reasoning tokens for decision tasks where there was a conflict between enjoyment and accuracy (see Figure~\ref{fig:reasoning-tokens} and S.I. Figure~\ref{fig:reasoning-tokens-full}). All $10$ consumed more tokens in the conflict zone when the stakes were low ($q < 0.001$) and in the high stakes setting ($q < 0.01$). Further, $9$ of $10$ models consumed relatively less tokens when the stakes were high compared to low ($q < 0.001$; Gemini-4 used more at high stakes). This is interesting, as it suggests that LLMs may think \textit{less} for some allocation decisions when the stakes are high than low. 

Human participants in our studies were required to spend at least five seconds before responding, complicating direct comparisons of deliberation time. Nonetheless, post-survey responses suggest participants were open to receiving assistance from external help for some questions (see S.I.~\ref{sec:post-survey}). This raises questions for future study around the cognitive demands of navigating these tradeoffs for people and how such demands may inform whether and when people engage decision support, as well as how expenses from LLM-based allocation may vary depending on the nature of the decision.

\section*{Discussion}

As AI systems are increasingly integrated into workplaces, managers will be faced with a question of who to delegate a task to: a human employee or AI system. This choice may come with a tradeoff -- whether to favor the expected outcomes of work (siding with the most capable agent) or weigh the deeply human experience of engaging in work, and the enjoyment (or lack thereof) that may come with it. 

Whether and how these tradeoffs are reconciled has substantial ramifications for the everyday experience of individual workers, as well as potential macro-effects, both economically and politically. Our work builds on previous investigations of the ``jagged'' integration of AI into human workplaces~\citep{dell2023navigating}. Specifically, our results suggest that this process will be impacted not only by the relative performance of workers, but (1) the human experience of work, (2) the direction of this experience (e.g., for tasks that would be more aversive to people), (3) the broader context (e.g., stakes), and (4) who or what is even making these choices in the first place (a human or LLM). This could mean, for instance, that people and LLMs favor assigning humans to lower stakes work that they enjoy, or higher stakes work where they outperform AI systems, even if it means the human worker does not enjoy that task. How these stakes are defined, and by whom, deeply matters, as is evidenced by broader discourse around AI's integration into human work and society~\citep{shao2025future, liang2024monitoring, epstein2023art, messeri2024illusions, wga2023mba, leiden2026declaration, brynjolfsson2025generative, liu2026medical, topol2026future}.

Our goal is not to make recommendations about how to navigate this tradeoff, but rather to highlight that this tradeoff exists and that people (and many LLMs) are sensitive to the balance of objective outcomes and subjective human experience in such allocation decisions. Indeed, our work is just a first step to characterize these tradeoffs in people and LLMs. Here we study participants recruited from the United States via a crowd-sourcing platform (see Methods). Future work should explore the generality of our findings to real managerial choices, as well as participants recruited cross-culturally, who may place different favor on subjective experience over objective reward ~\citep{henrich2001search} and otherwise have different preferences for human versus AI work~\citep{ge2024culture, awad2018moral, gillespie2025trust}. Similarly, future work can better understand the sensitivity of different LLMs to these tradeoffs under a wider range of prompts and response modes. We see our setup as a microcosm that can be used to study value alignment of LLMs, which might be especially useful for companies testing whether and where LLMs can be deployed reliably and where they may differ in practice when there are tradeoffs between outcomes and more subjective measures like enjoyment.

There are a wide range of naturalistic modifications that can be built on top of our paradigm as well. We assumed that the manager already knows the relative expected accuracy of both agents and expected enjoyment of a task. While estimates of accuracy might be sensible in some contexts (e.g., particularly with more fine-grained evaluations of expected AI performance ~\citep{zhou2026general, wang2026well}), in many complex real work contexts, how well people or AI may do on a task may not be known, much less how enjoyable that task might be to the human worker -- and it is similarly unclear whether LLMs would align with people in ex ante estimates of task outcomes or human enjoyment~\citep{collins2026evaluating}. Although we informed the task allocators that the expected accuracy already incorporates the expected effort that a worker may put in, enjoyment in work may substantially influence the outcomes~\citep{graves2012driven, hackman1976motivation}. Allocators were also informed that each task was done by a different random worker. In practice, decisions may be made over time, with the same worker, and one might imagine that risk of attrition influences weight on enjoyment. Allocation decisions might also be influenced by the broader competitive context and could include other compensatory measures like wage adjustment if assigning tasks that impact experience~\citep{korinek2022preparing, brynjolfsson2025research, gans2026replaceable, dong2025experimental}. Here, we also assumed that the human worker is always doing \textit{something} at work (their baseline task), however, choices might not be between two kinds of labor but between labor and leisure with different degrees of meaningfulness~\citep{kool2014labor, campbell2025effortful, csikszentmihalyi1989optimal} and other sources of meaning, e.g., volunteering~\citep{meier2008volunteering, chan2026work}. Future work can better understand whether, and if so why, people and LLMs may be sensitive to such factors -- and not only sensitive in choices but in the effort underlying such choices that might inform when people also want to defer the choice of a decision~\citep{sunstein2023decisions} (see S.I. Section~\ref{sec:post-survey}) and how this relates to their intuitions about leadership~\citep{edelson2018computational}, team dynamics~\citep{mieczkowski2025people, xiang2023collaborative, mckee2024warmth}, and the function of managers more generally~\citep{athey1994allocation, goldberg2026purpose, lazear2015value, simon1947administrative}.

That we see people weighing objective and subjective experience in allocation choices at all, even in our current simple setup, motivates wider incorporation of other kinds of subjective factors into models of human decision making. While ample studies in the behavioral sciences, from economics to behavioral game theory and cognitive science, have explored how people account for factors beyond productivity, like fairness, when making decisions that involve others ~\citep{rabin1993incorporating, fehr1999theory, charness2002understanding, dimick2018models, gates2020helpful, kleiman2025evolving, chandra2026computational, li2026large, marjieh2024task}, comparatively less work has investigated -- much less integrated into formal models -- the ways that people account for other subjective \textit{experiential} factors like enjoyment. A growing body of work has begun to build and empirically test finer-grained computational models of how people reason about the experience of ``funness''~\citep{brandle2025leveling, chu2025makes, collins2026people, li2021expectations}, more aversive experiences, like pain~\citep{berke2026toward}, regret ~\citep{houlihan2023emotion, chen2024intervening, collins2025empathy, petrova2025learning}, or general unhappiness~\citep{dubey2022pursuit}; yet, whether and how to integrate such richer models of human behavior into more classical decision-theoretic frameworks that might inform policy remains an open question. Our results call further attention to this gap and the necessity of better understanding how people and AI systems may assess and integrate notions of human experience into decisions~\citep{cai2023incentive}.

As AI is increasingly integrated into human workplaces, we urgently need to understand decision-maker's intuitions -- whether they be people or LLMs -- for how to navigate the inevitable tradeoffs that arise from such integration, such as which lead people to weigh others' subjective experience beyond just the expected outcome. How much we enjoy our work -- now and in the future -- depends on it.

\section*{Materials and Methods}

\subsection*{Stimuli}

For all experiments, respondents (participants or LLMs) played the role of a manager, allocating tasks between a human worker or AI system. Tasks to be allocated were framed as ``special projects'' that, if assigned to a human, the human worker would do instead of their ``baseline'' task. Each task was presented with the expected difference in capability (accuracy) that the human worker versus AI was expected to obtain if assigned that task. Additionally, tasks were presented with the expected relative enjoyment of that task for the human worker over the human's baseline task. Accuracy differences could range from $-100$ to $100$ and were presented as expected human accuracy minus expected AI accuracy. Enjoyment levels could also range from $-100$ to $100$, where $-100$ meant that the task was ``much less enjoyable than the baseline task'' and $100$ was ``much more enjoyable than the baseline task.'' 

We designed $315$ tasks which spanned the full range of enjoyment levels (in steps of $10$ between $-100$ and $100$) and were restricted to relative differences in accuracy in the range between $-35$ and $35$ in steps of $5$, based on pilot data suggesting this range was the more interesting region to assess potential tradeoffs between expected performance and worker experience. For the stakes condition, each of the $315$ tasks was also labeled with each stakes level (low, medium, or high), yielding $945$ tasks.

\subsection*{Human experiments}

\paragraph{No-stakes setting.}

Participants ($N=344$) were recruited via Prolific to participate in the ``no-stakes'' task allocation setting. Participants were recruited from the United States and required to speak English. The study was expected to take approximately $15-20$ minutes and participants were paid at a rate of $\$12$/hr. Participants were required to use a laptop. Participants provided informed consent. The study was conducted under the Princeton University IRB. 

The $315$ tasks were split into five batches of $63$ tasks. Each participant made allocation decisions for $63$ tasks. Expected relative accuracy and enjoyment were presented on slider scales with the number above the slider. For each task, participants made their choice of who to allocate to (human or AI worker) by clicking on a button; participants were only allowed to select one. Button order was shuffled across participants. Participants were required to spend at at least five seconds on the screen before they could make their selection, but they were allowed to take as much time as they wanted. At the end of the study, participants provided a few open-ended post-survey questions (see S.I.~\ref{sec:post-survey}).

Our pre-registered quality control was intentionally strict to guard against potential bots. Specifically, we ran custom bot and attention detection software tracking mouse movements as well as whether a participant’s window was blurred (indicating losing focus or leaving the tab). We also had a hidden checkbox that discouraged AI agents from participating in the study. As the interface was set up such that participants need to click who to allocate to, then press submit (with some separation between the two), we expected substantive mouse movement between the two. We excluded participants who: (1) checked the hidden checkbox, (2) failed the comprehension quiz three or more times, (3) spent one or more minutes outside of the main window, as detected by our window-blur tracker, (4) had five or more instances of window blurring, and (5) had an average of at most two mouse movements/click over the allocation trials. These thresholds were based on pilot data. Based on our pre-registered criteria, collection continued until there were $40$ participants passing quality control for each batch. This led to $200$ participants' allocation choices being kept ($107$ failed our quality control criteria; we also removed $17$ participants who attempted to take multiple studies across our experimental conditions; $20$ additional participants were not included due to over-recruitment). 

\paragraph{Stakes.}

We ran a separate study on Prolific ($N=905$) wherein participants engaged in the same task but saw tasks labeled as either low, medium, or high stakes. Participants again made allocation decisions for $63$ tasks; $21$ of the tasks were labeled as low, $21$ as medium, and $21$ high stakes, respectively. We divided the set of $945$ tasks into $15$ batches and based on our pre-registration, continued collection until $40$ participants passed quality control for each batch. This produced $600$ valid participant judgments ($275$ failed our quality control, where we applied the same quality control as described above, and $22$ were dropped for having repeated multiple of our studies; $8$ additional were dropped from overrecruitment). All other details on recruitment followed the no-stakes setting.

\subsection*{LLM allocations}

We tested a suite of large language models (LLMs) on the same allocation choices. LLMs were prompted with a lightly modified version of the instructions provided to participants (see S.I.~\ref{sec:si-llm-prompts}). LLMs were presented tasks independently and sampled $40$ times for each task. Temperature was set at $0.5$ for all models that permitted it (unspecified for OpenAI reasoning models, Claude Opus, and the Grok 4 models). For LLMs with a possible reasoning mode, the default reasoning level was used. All LLMs were queried through APIs. Closed-sourced models were queried through their own API; open-source models were queried through the third-party OpenRouter platform.

\subsection*{Analysis methods} 

Our primary analyses were conducted in R (v4.5.1) using \texttt{glm}, \texttt{lmtest}, and \texttt{sandwich}, with most plots using \texttt{ggplot2}.

\paragraph{Notation.} Each trial (allocation decision) $j$ was tagged with the expected difference in accuracy between the human worker and AI system ($\Delta x_j = x_{h,j} - x_{\text{AI},j}$) and the human worker's expected enjoyment of the task relative to their baseline task ($\rho_{h,j}$). Let $y_{ij} \in \{0, 1\}$ indicate whether the respondent $i$ assigned trial $j$ to the human worker ($y_{ij} = 1$) or AI system ($y_{ij} = 0$). In the stakes experiments, let $s_j \in \{\text{low}, \text{med}, \text{high}\}$ denote the stakes level of trial $j$.

\paragraph{Logistic regression model.} As described in our pre-registration, our primary question was whether people, in aggregate, weigh human enjoyment ($\rho_h$), beyond just the difference in agent capabilities ($\Delta$), when allocating tasks between people and AI systems. In the no-stakes condition, we tested whether accuracy difference and enjoyment each matter by fitting a logistic regression model to allocation choices: 
\begin{equation}
    \Pr(\text{assign human}) = \Pr(y_{ij} = 1) = \sigma\!\left(\beta_{\text{offset}} + \beta_{\text{acc}}\,\Delta x_j + \beta_{\text{enj}}\,\rho_{h,j}\right).
    \label{eq:main-model}
\end{equation}

The model was fit on the choices pooled across participants per trial, with clustered standard errors by participants. We hypothesized coefficients ($\beta_{\text{acc}}$ and $\beta_{\text{enj}}$) will each be positive. We took a coefficient that is positive with $p < 0.05$, under a Wald $t$-test, as support for the hypothesis that that feature ($x_h - x_{\text{AI}}$ and/or $\rho_h$) mattered. We then confirmed each feature contributes over the other using participant-cluster-robust Wald tests,  comparing the full model in Equation ~\ref{eq:main-model} to models with only $\Delta x$, only $\rho_h$, or only an intercept, respectively.

For the stakes experiment, we fit a joint model wherein the intercept and both weights vary by stakes level:
\begin{equation}
    \Pr(y_{ij} = 1) = \sigma\!\left(\beta_{\text{offset}}^{(s_j)} + \beta_{\text{acc}}^{(s_j)}\,\Delta x_j + \beta_{\text{enj}}^{(s_j)}\,\rho_{h,j}\right),
    \label{eq:stakes-model}
\end{equation}
implemented as a model with stakes level as a categorical variable interacting with $\Delta x$ and $\rho$, with low stakes as the reference level. As in the no-stakes case, the model was fit on pooled participant allocations with clustered standard errors by participants. We assessed whether the weights depend on stakes with a participant-cluster-robust Wald test comparing Eq.~\ref{eq:stakes-model} to a model holding $\beta_{\text{acc}}$ and $\beta_{\text{enj}}$ fixed across stakes levels. We predicted that increased stakes would increase the weight on relative agent accuracy ($\beta_{\text{acc}}^{(s)} - \beta_{\text{acc}}^{(\text{low})} > 0$) and decrease the weight on enjoyment ($\beta_{\text{enj}}^{(s)} - \beta_{\text{enj}}^{(\text{low})} < 0$) for $s \in \{\text{med}, \text{high}\}$. We ran participant cluster robust $t$-tests on the corresponding interaction terms.

We also predicted that stakes change how participants’ weighting of agent capabilities and enjoyment relate to one another. To test, we conducted a likelihood ratio test between a model that accounts for three-way interactions (i.e., $\Delta x \times \rho_h \times \text{stakes}$ interactions) versus one that only considers pairwise interactions between $\Delta x$, $\rho_h$, and stakes level. We took $p < 0.05$ as support for three-way interactions between stakes, $\Delta x$, and $\rho_h$.

\paragraph{Asymmetry.} To assess potential asymmetry in how enjoyment is weighed, we also fit a piecewise logistic model with separate slopes for $\rho_h > 0$ and $\rho_h < 0$:
\begin{equation}
    \Pr(y_{ij} = 1) = \sigma\!\left(\beta_{\text{offset}} + \beta_{\text{acc}}\,\Delta x _j + \beta_{\text{enj}}^{-}\min(\rho_{h,j}, 0) + \beta_{\text{enj}}^{+}\max(\rho_{h,j}, 0)\right),
    \label{eq:piecewise-model}
\end{equation}
where $\beta_{\text{enj}}^{-}$ is the weight on enjoyment in the region where the human worker would enjoy the special project less than their baseline task ($\rho_h < 0$), and $\beta_{\text{enj}}^{+}$ is the weight in the region where they would enjoy it more ($\rho > 0$). We tested for asymmetry with a participant-clustered Wald $t$-test on $\beta_{\text{enj}}^{+} - \beta_{\text{enj}}^{-}$, as well as a test on the bootstrapped difference in exchange rates between the two regions (see below). In the stakes experiment, the piecewise model was fit separately to the choices made at each stakes level.

\paragraph{Exchange rate.}

For each setting, we also analyzed the ``exchange rate'' (ER), how many points of enjoyment are needed for a shift in accuracy difference between the workers. Exchange rates were computed from the coefficients for both the main and piecewise logistic models: 
\begin{equation}
    \text{ER} = \frac{\beta_{\text{acc}}}{\beta_{\text{enj}}}.
    \label{eq:exchange-rate}
\end{equation}

We estimated uncertainty in the exchange rate using non-parameterically resampled nonparameteric bootstraps, bootstrapped over participants. We report $95\%$ confidence intervals over $10,000$ bootstraps. 

In the stakes setting, we used exchange rate to more specifically test our prediction that relative weighting of enjoyment decreases with increased stakes. Specifically, we tested whether ER differed across stake levels ($\text{ER}^{s}$ for stakes level $s$). We ran $10,000$ bootstrap samples fitting a model for each stake level, resampling over participants. As specified in our pre-registration, we hypothesized that exchange rate will be significantly lower for the low stakes trials than medium and high stakes and that the exchange rate will be significantly lower for the medium than high stakes. We ran pairwise comparisons taking the difference between each bootstrapped sample (i.e., med $-$ low; high $-$ low; high $-$ med) and computing the number of bootstrapped samples where the difference is $<= 0$ (against our predicted direction). We took $p < 0.05$ as support for the hypothesis that the exchange rate increases for higher stakes within each comparison.

\paragraph{LLM comparisons.} We ran similar analyses for each LLM, fitting the same logistic models to the number of samples $k_{m,j}$ (out of $40$ rollouts) in which LLM $m$ assigned trial $j$ to the human worker, using a binomial likelihood. To estimate uncertainty in the exchange rates, we bootstrapped over $40$ responses per LLM per trial (sampled with replacement and recomputing $\hat{k}_{m,j}$) and refitting the model. Note that this is similar to our bootstrapping over participants in the human data; however, in the human data, each of the $40$ responses was a different person (and we cluster and bootstrap based on each person's set of judgments). With each LLM, the rollouts are treated as unrelated across trials.

We assessed differences between each LLM's judgments and people's through several exploratory statistical analyses. While we report ER, to compare exchange rates in a way that avoids possible blowup in values (e.g., if $\beta_\text{enj} \approx 0$), we run bootstrapped comparisons between people and LLMs using the inverted $w = 1/\text{ER}$. We thereby took the difference between bootstrap draws (participant bootstraps for people; response bootstraps for LLMs) on $w$ and assessed a two-sided p value on the difference. We separately compared the dynamic range (assessing $\text{ER}^\text{high} - \text{ER}^\text{low}$) for each setting using bootstrap fits on the models and people. For all analyses, we used a Benjamini-Hochberg correction due to multiple testing hypotheses (reported as $q$).

Note, at high stakes, five LLMs (DeepSeek-V4-Pro, DeepSeek-V4-Flash, GPT-6.1 Sol, Grok 4.20 reasoning, Llama-3.3-70B) always sided with the more accurate worker whenever there was a conflict between accuracy and enjoyment. For DeepSeek-V4-Pro GPT-6.1 Sol, and Grok 4.20 (reasoning), ERs were unbounded as rollouts produced identical judgments on nearly all trials, leading the logistic fit to not converge; this did not occur with the other two models (DeepSeek-v4-Flash and LLaMA-3.3-70B) as there was sufficient variability in some rollouts for non-conflict trials. We count the three unconverged models as still having a higher-than-human high-stakes ER and a wider dynamic range given they always assigned to accuracy over enjoyment in conflict regions. 

\paragraph{Reasoning token usage.} We conducted exploratory analyses on the subset of LLMs where reasoning token usage was trackable. We assessed the relative token usage when there was a conflict between human subjective experience and accuracy (i.e., when $\Delta$ and $\rho_h$ have opposite signs), as well as how token usage differed across experimental conditions (i.e., the different stakes level and no-stakes level). As different LLMs expend different amounts of reasoning tokens, we normalized the the token usage within each LLM type across the experiments. That is, we computed the median reasoning token amount over all decisions made by a given LLM $m$ (pooled across all trials and stake settings). For each individual trial, we then took the median reasoning token usage for that trial (over the $40$ rollouts from the LLM) and divided by the median pooled reasoning token amount. For people, we applied a similar procedure to their  response times, normalizing based on the median pooled response time (minus the required minimum five seconds per trial). 

To understand whether reasoning token amount differed in the conflict regions, in exploratory analyses, we then computed a Wilcoxon ranked-sum test on the conflict versus non-conflict zones within each stakes setting, as well Wilcoxon signed-rank tests between the low and high stakes paired sets of tasks. We again applied a Benjamini-Hochberg multiple hypothesis correction for each test.

\subsection*{Data and code availability} 

All code and anonymized data will be made available upon publication.

\subsection*{AI use statement} 

We used AI for help coding the statistical analyses, human experiments, LLM queries for task allocation and post-survey response categorization, and data visualization including some of the plots and tables for the manuscript. We used AI to help with manuscript formatting, providing suggestions for reporting statistics, and surfacing additional related work we might consider; otherwise, AI was not used in writing. We take responsibility for any errors.

\subsection*{Acknowledgments} 

We thank Lio Wong, Judy Fan, Umang Bhatt, Jose Hernandez-Orallo, Scott Stern, Nori Jacoby, Haoyu Hu, Chenyi Li, Graham Todd, Stephen Cave, Eric Horvitz, Neil Lawrence, Lucy Hampton, and David Autor for valuable conversations that informed this work. We also thank Jackie Lane, Simon Friss, and the Laboratory for Innovation Science at Harvard; Hita Kambhamettu and Andrew Head and the Penn HCI group; the Princeton Computational Cognitive Science Group; and the Purdue AI Thought Partner Lab for helpful feedback and questions on early presentations of this work. We thank Kevin Smith for help developing bot detection in our human experiments. KMC acknowledges support from the NSF SBE SPRF. MKW acknowledges supported by Cooperative AI Foundation, Toyota Research Institute (TRI), UW-Tsukuba Amazon NVIDIA Cross Pacific AI Initiative, Templeton World Charity Foundation (https://doi.org/10.54224/34843). This work was supported by the Office of Naval Research (ONR) MURI grant N00014-22-1-2813, DARPA cooperative agreement HR0011252E002, and the Air Force Office of Scientific Research under award number FA9550-23-F-0014 to EM.

\bibliography{main}
\bibliographystyle{abbrvnat}

\newpage

\appendix
\tableofcontents

\section{Additional human experiment analyses}

\begin{figure}[h!]
    \centering
    \includegraphics[width=1.0\linewidth]{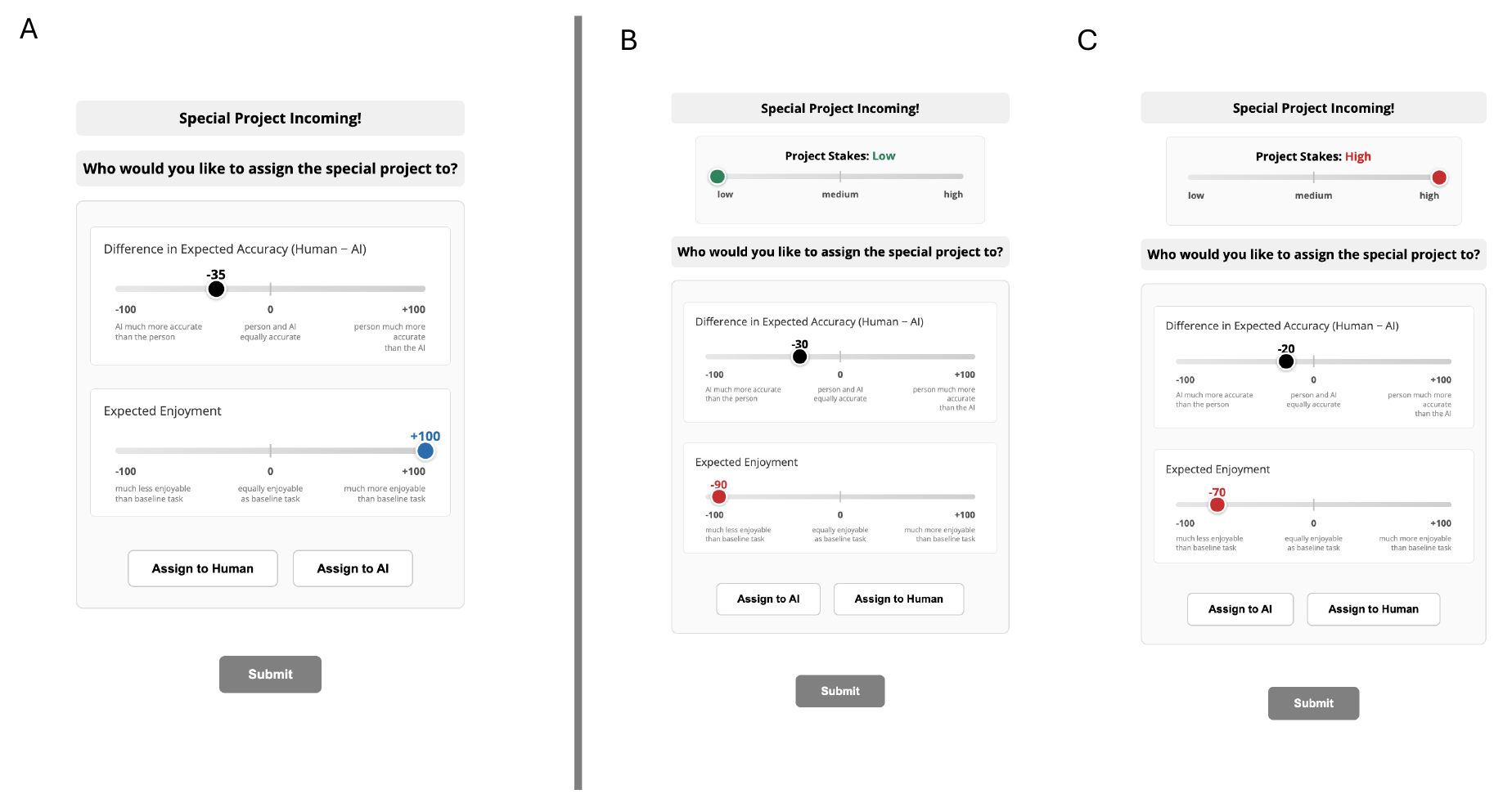}
    \caption{\textbf{Example experiment interfaces.} \textbf{(A)} Example interface human participants saw for the no-stakes experiment; \textbf{(B)-(C)} example interfaces participants saw when stakes were specified shown for one example low (B) and high (C) stake settings.}
    \label{fig:experiment-interface}
\end{figure}

We include additional analyses into the human data, as well as screenshots of the main experimental trials (Figure~\ref{fig:experiment-interface}) as well as full instructions in Section~\ref{sec:human-instructions}. 

\subsection{Asymmetry in enjoyment weight}

\begin{figure}
    \centering
    \includegraphics[width=1.0\linewidth]{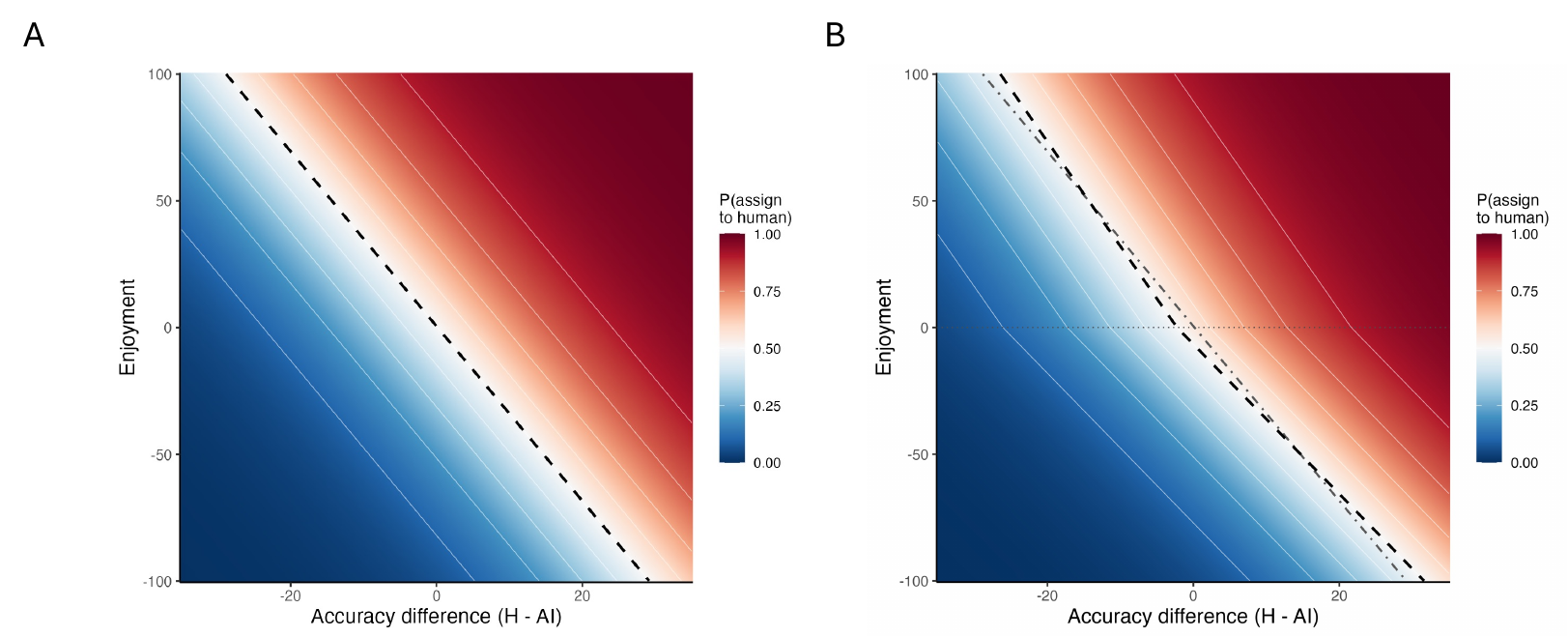}
    \caption{\textbf{Contours of logistic regression models fit to human choices in no-stakes setting.} \textbf{(A)} Contour plot with single parameter for human experience compared to \textbf{(B)} a piecewise model which fits a separate parameter for $p_h>0$ and $p_h<0$, revealing an asymmetry in how people navigate the tradeoff.} 
    \label{fig:contour-plots-no-stakes}
\end{figure}

We include contour plots of the logistic models fit to human choices in the no-stakes setting, comparing a model with a single slope fit (Figure~\ref{fig:contour-plots-no-stakes}A) to enjoyment versus a piecewise model to capture asymmetry (Figure~\ref{fig:contour-plots-no-stakes}B), demonstrating the stronger weight of human subjective experience when that experience is framed as aversive (less enjoyable) than positive (more enjoyable). 

\begin{figure}
    \centering
    \includegraphics[width=1.0\linewidth]{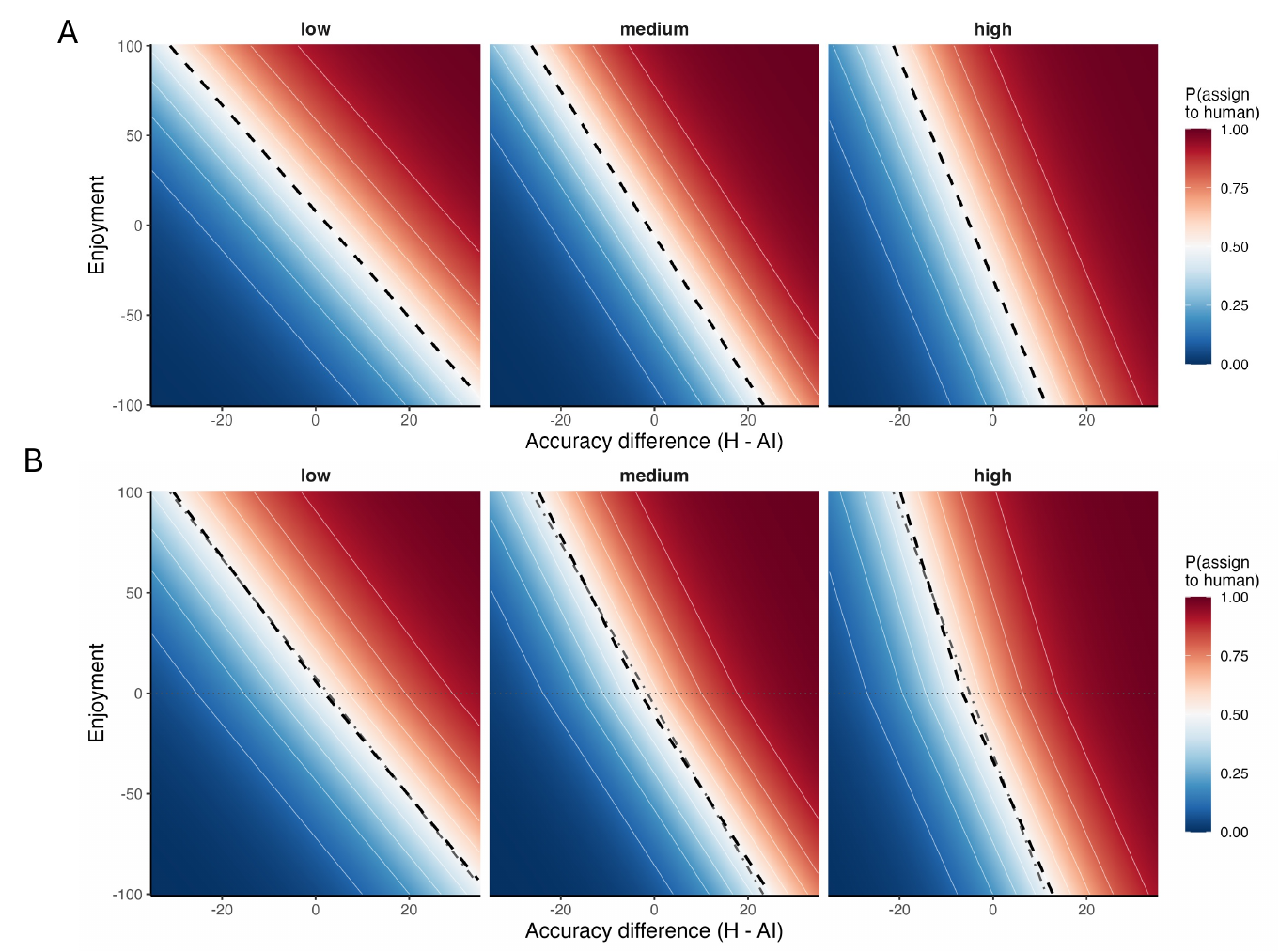}
    \caption{\textbf{Contours of logistic regression models fit to human choices in stakes.} \textbf{(A)} Contour plot with single parameter for human experience compared to \textbf{(B)} a piecewise model which fits a separate parameter for $p_h>0$ and $p_h<0$, revealing an asymmetry in how people navigate the tradeoff. Models are fit separately to choices made for each stakes setting.} 
    \label{fig:contour-plots-stakes}
\end{figure}

\begin{figure}
    \centering
    \includegraphics[width=0.5\linewidth]{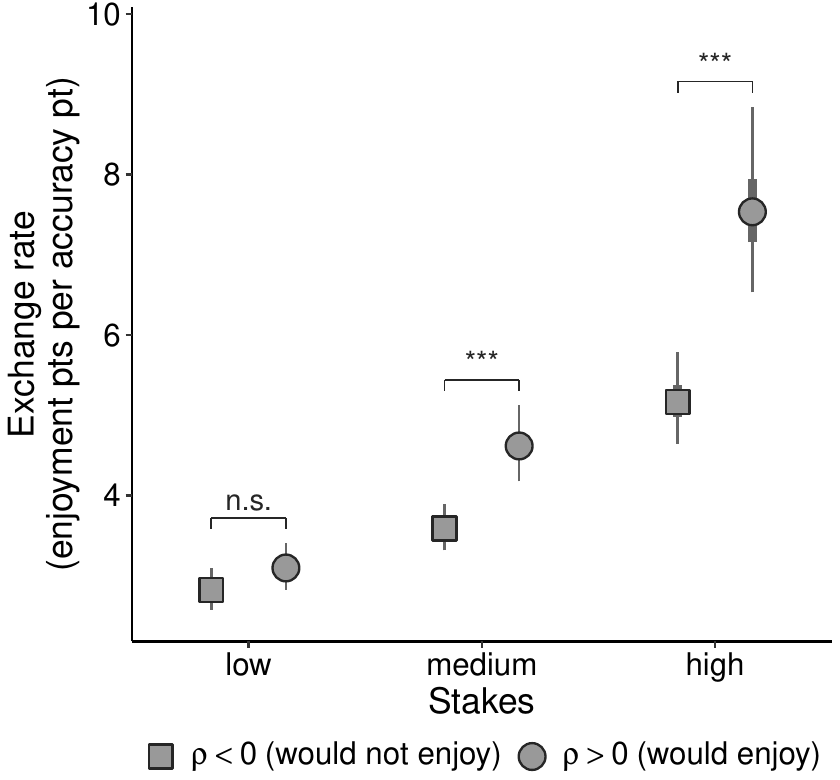}
    \caption{\textbf{Exchange rate asymmetry over stakes.} Exchange rate (ER; relative weighting of accuracy over enjoyment) for models with a kink at $p_h > 0$ and $p_h <0$ by stakes. ER increases as a function of stakes, and in the medium and high stakes conditions, involves a lower relative weight of accuracy over enjoyment when the task allocation would lead to less (rather than more) human enjoyment.} 
    \label{fig:exchange-rate-asymm-stakes}
\end{figure}

Similar asymmetry appears in the medium and high stakes settings, but not low stakes, as evidenced in both the contour plots (see Figure ~\ref{fig:contour-plots-stakes}) and exchange rates (see Figure ~\ref{fig:exchange-rate-asymm-stakes}).



\subsection{Post-survey questionnaire analyses}
\label{sec:post-survey}

\begin{figure}
    \centering
    \includegraphics[width=1.0\linewidth]{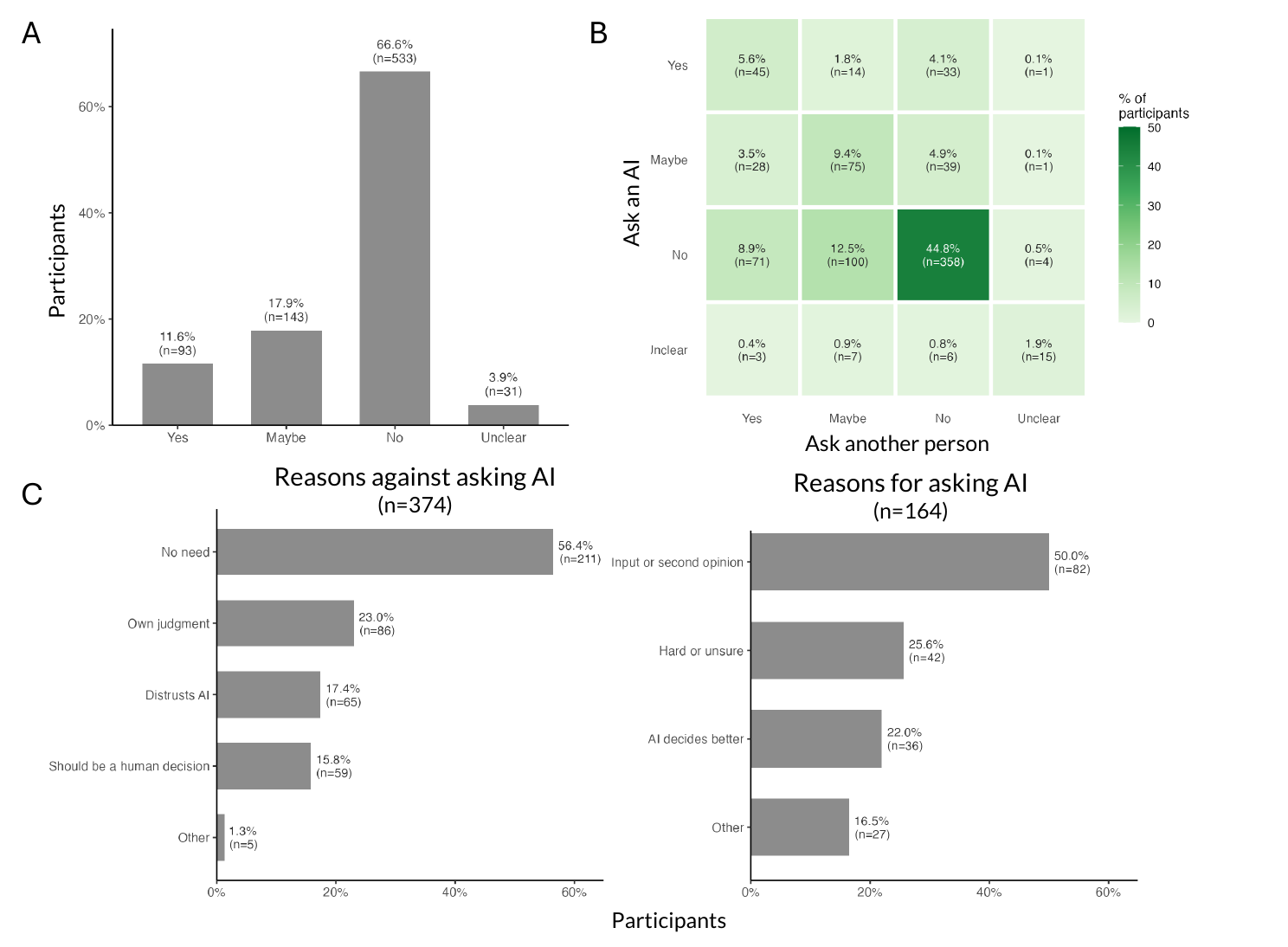}
    \caption{\textbf{Categorizing participant post-survey responses on potential interest (or lackthereof) in accessing AI for task allocation input.} \textbf{(A)} Percent of participants whose post-survey response were categorized as indicating they would be open to accessing AI for help in task allocation decisions. Participants are aggregated over the no-stakes and stakes conditions (i.e., $n=800$).; \textbf{(B)} Participants' responses broken down by whether they would be open to having AI input (y axis) versus human input (x axis). Color indicates percent of participants indicating that response.; \textbf{(C)} Automated categorization of the reasons participants provided (when they were coded as having provided at least one reason) for why they would or would not ask AI for help for task allocation choices.}
    \label{fig:post-survey-resps}
\end{figure}

Participants also answered several open-ended questions after their trials (see full listing in Section~\ref{sec:human-instructions}). In particular, participants indicated whether they would have been inclined to have help from either an AI system and/or other person for any of their allocation decisions. 

In exploratory analyses, we categorized participants' responses to assess how many participants were keen for AI versus external human assistance in any of the allocation decisions and what rationales they provided for their responses. We used a set of LLMs to categorize the responses. The categories were constructed by one author from our team who read through many of the responses. The prompt (see Section~\ref{sec:categorization-prompt}) was co-constructed between the author and Claude. For each participant response, we took three LLMs (Claude Opus 5.5, DeepSeek-v4-Pro (with reasoning enabled), GPT-5) and as the LLM coding is stochastic and all participant responses are in open-ended natural language (which has some ambiguity), we had each LLM categorize each response three times. From this, we took the majority vote over the responses, and then took the majority over the three LLMs. The codings should be as suggestive but not definitive. 

The majority of participants indicated no inclination for AI assistance in their response (Figure ~\ref{fig:post-survey-resps}A). A moderate portion (approximately $20\%$) of participants were potentially open to some human assistance but no AI assistance (see Figure ~\ref{fig:post-survey-resps}B). The dominant reasons for not wanting AI assistance, when specified, was that the person found the task simple enough without need for AI or that they felt it was a decision that should be made on their own (Figure~\ref{fig:post-survey-resps}C). Of the participants who did indicate a reason for accessing AI, the dominant reason provided was generally that they wanted a second opinion or other kind of input from the AI to help them make their choice (Figure~\ref{fig:post-survey-resps}C).

\subsection{Response time exploratory analyses} 

\begin{figure}
    \centering
    \includegraphics[width=1.0\linewidth]{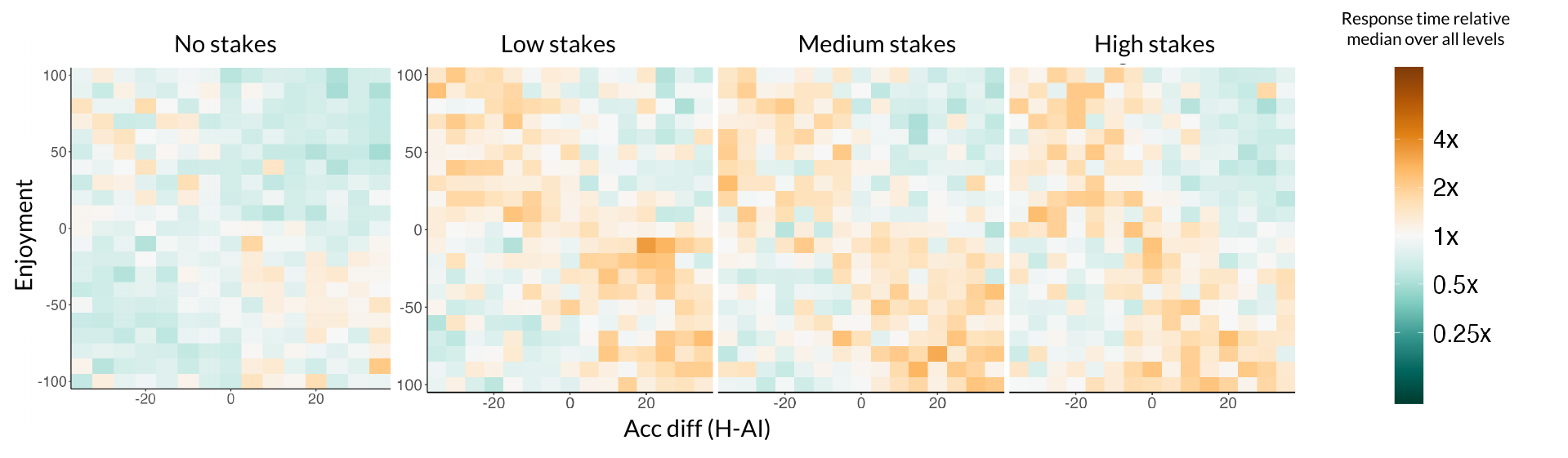}
    \caption{\textbf{Normalized response time for participants.} Participant response times, normalized by response times over all trials and levels. Darker indicates more relative time spent on any one trial. However, note participants were required to spend at minimum $5$ seconds on any given trial.}
    \label{fig:rxn-time}
\end{figure}

\begin{figure}
    \centering
    \includegraphics[width=1.0\linewidth]{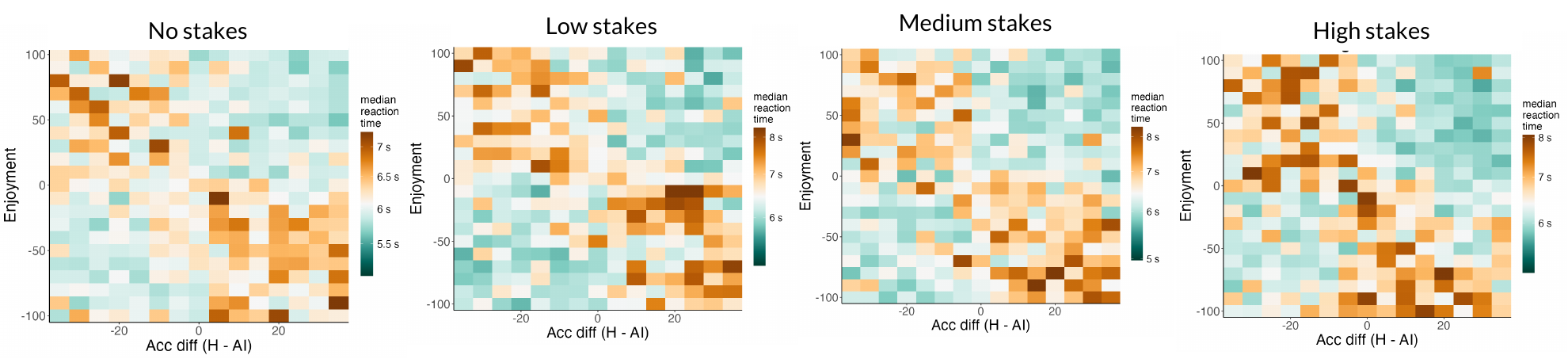}
    \caption{\textbf{Raw response time for participants.} Raw, unnormalized participant response times (in seconds). Note, the scales change across settings.}
    \label{fig:rxn-time-raw}
\end{figure}

We additionally explore participant response times as a function of trial properties and stakes (Figure~\ref{fig:rxn-time}). We apply a similar normalization procedure as to the LLM reasoning token analyses, but now, over response time (in seconds). That is, we compute the median response time aggregated over all participants, all trials, and all experiment settings (e.g., different stake levels). Any given response time is then normalized based on the median aggregated time. Figure~\ref{fig:rxn-time-raw} depicts unnormalized response times. We note caution, however, in directly reading the reaction times as participants always had to spend at least $5$ seconds on their trial before making a decision. It is therefore possible participants were inclined to respond sooner for some trials. 

\subsection{Robustness to trial presentation}
\label{si:just-slider}

\begin{figure}
    \centering
    \includegraphics[width=0.8\linewidth]{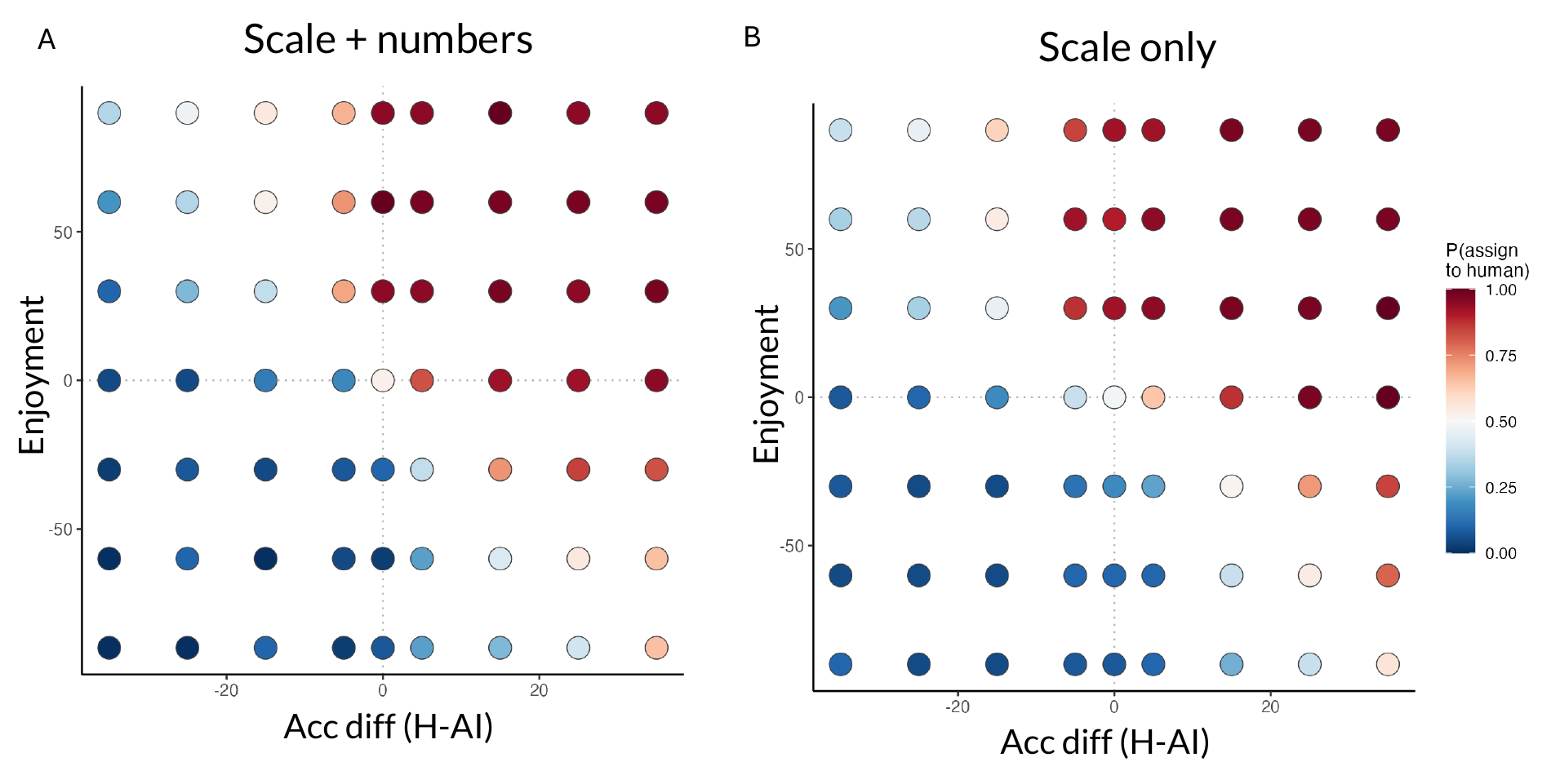}
    \caption{\textbf{Replication where trials are presented with only visual scales, no numbers.} \textbf{(A)} The subset of trials from the full no-stakes experiment, where trials included scales and numbers presenting accuracy difference and relative enjoyment, compared against \textbf{(B)} the proportion of participants assigning trials to the human worker in the control wherein the raw numbers are not included.}
    \label{fig:just-slider}
\end{figure}

Participants in our main human studies saw the expected performance and enjoyment of workers on a slider scale with the actual number presented above the slider. To assess the influence of the numerical presentation, we ran a controlled follow-up study on one batch of $63$ tasks from our full $315$ stimuli space (i.e., tasks were sampled from the range $-90$ to $90$ enjoyment in steps of $30$ and accuracy differences between $-35$ and $35$ in steps of $10$, including the additional accuracy difference of $0$ level; see Figure~\ref{fig:just-slider}). We recruited participants under the same settings described in the main experiments, however, participants were shown the tasks only with the sliders not raw numbers of expected performance nor enjoyment. We recruited participants $56$ participants of which $39$ passed our quality control ($8$ failed quality control and $9$ were excluded due to having participated in multiple of our experiments). 

We compared the judgments made by participants in this control (scale only/no numbers) to judgments made by participants in the no-stakes condition who saw the scale and numbers (i.e., the $40$ responses per trial made for the trials that were included in scale only/no numbers control. Comparing models fit to judgments made by participants in each conditions revealed no significant difference in the tradeoffs (see Figure~\ref{fig:just-slider} and Table~\ref{tab:scale-control}). Note, we observe some qualitative difference at the $\pm 5$ accuracy difference, might have been hard to see that the dot was off the center without numbers. Future work can explore other stimuli presentations, e.g., judgments made over natural language descriptions of the task and workers. 

\input{figures/supp/scale_control_table}

\section{Additional LLM analyses}
\label{sec:si-llm-full}

\begin{figure}
    \centering
    \includegraphics[width=1.0\linewidth]{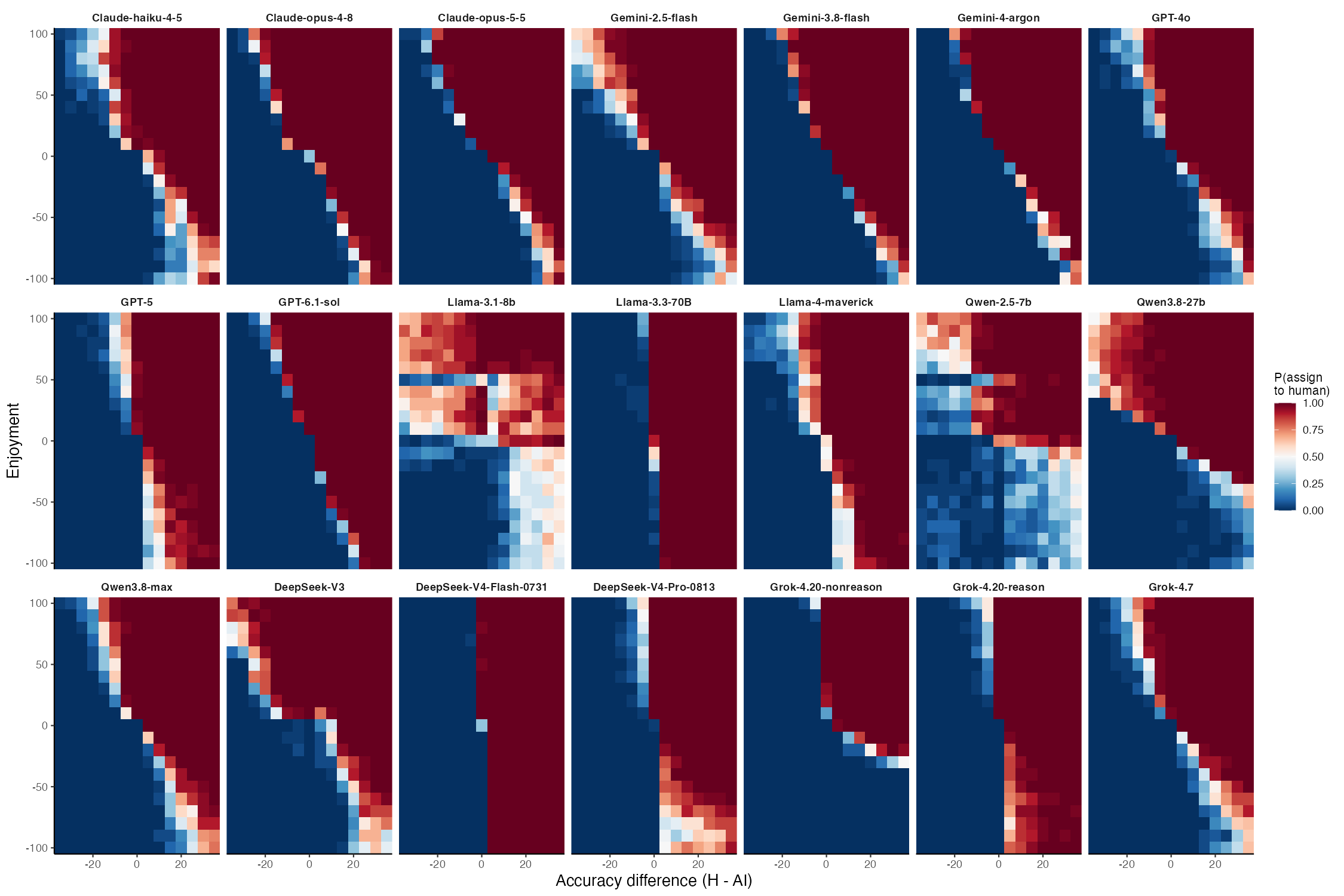}
    \caption{\textbf{LLM allocations, unspecified stakes condition}. Proportion of rollouts assigning the task to the human worker, for each LLM, in the no-stakes condition.}
    \label{fig:alloc-llm-no-stakes}
\end{figure}

\begin{figure}
    \centering
    \includegraphics[width=1.0\linewidth]{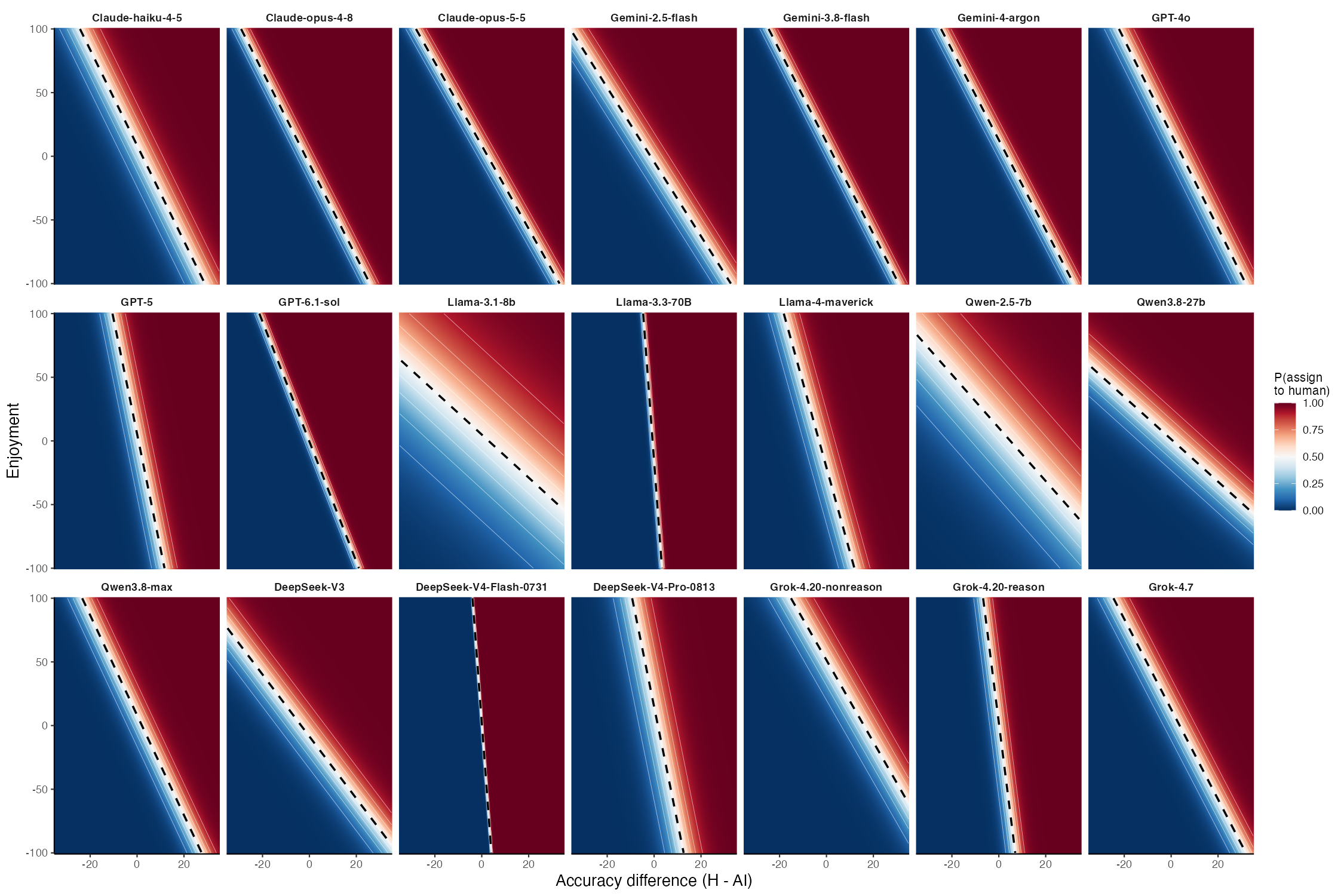}
     \caption{\textbf{Contours for logistic models fit to allocations in the unspecified (no-stakes) condition}. Models fit to LLM judgments, fit with a parameter on accuracy difference and human worker experience.}
    \label{fig:contour-llm-no-stakes}
\end{figure}

\begin{figure}[h!]
    \centering
    \includegraphics[width=1.0\linewidth]{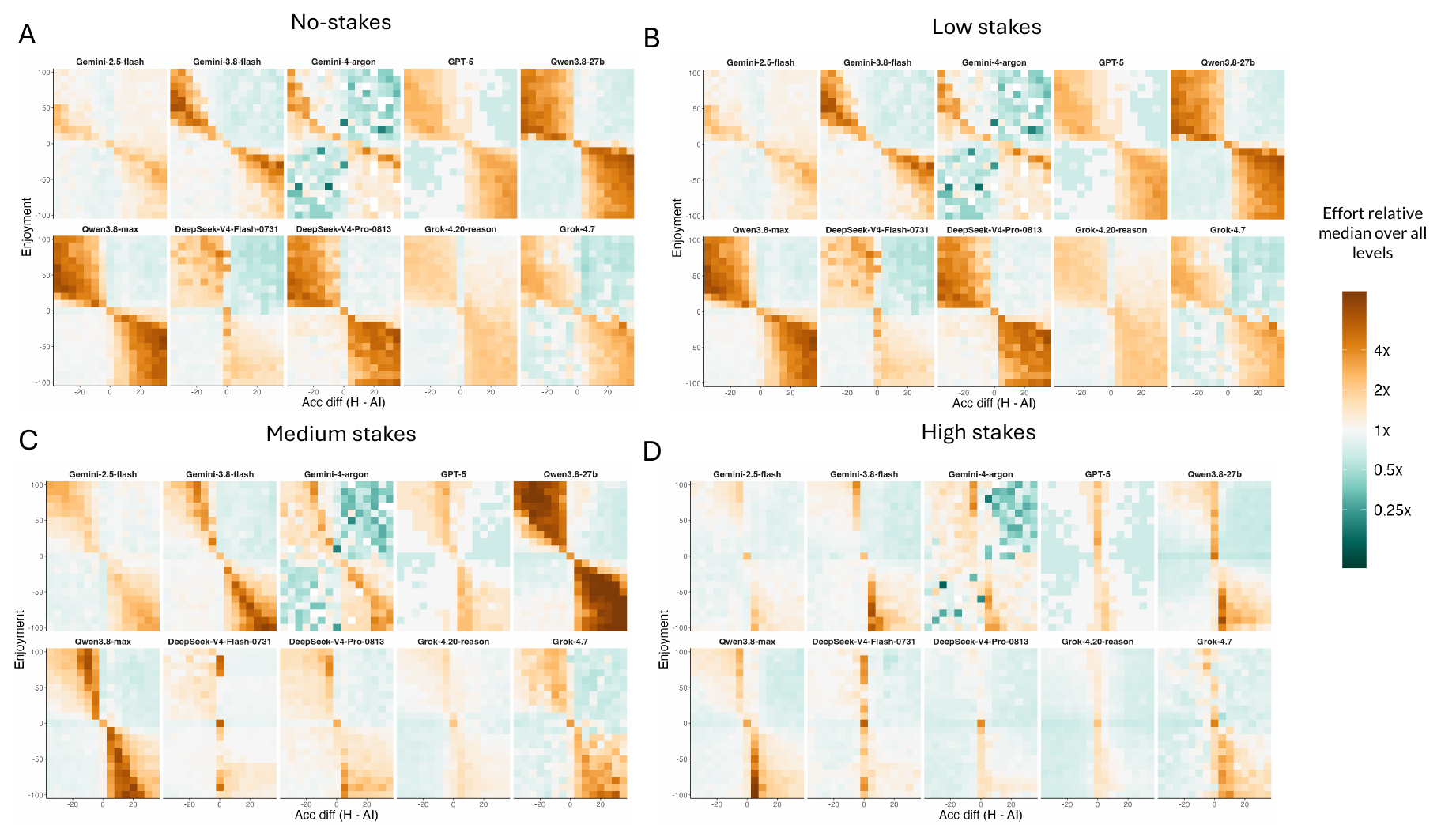}
    \caption{\textbf{Full Reasoning token consumption across allocation decisions}. Relative token consumption used by different LLMs across the trials. Token consumptions is normalized within each LLM and across all allocation decisions. The low and high stakes settings are reproduced here from the main text for ease of comparison.}
    \label{fig:reasoning-tokens-full}
\end{figure}

\begin{figure}
    \centering
    \includegraphics[width=1.0\linewidth]{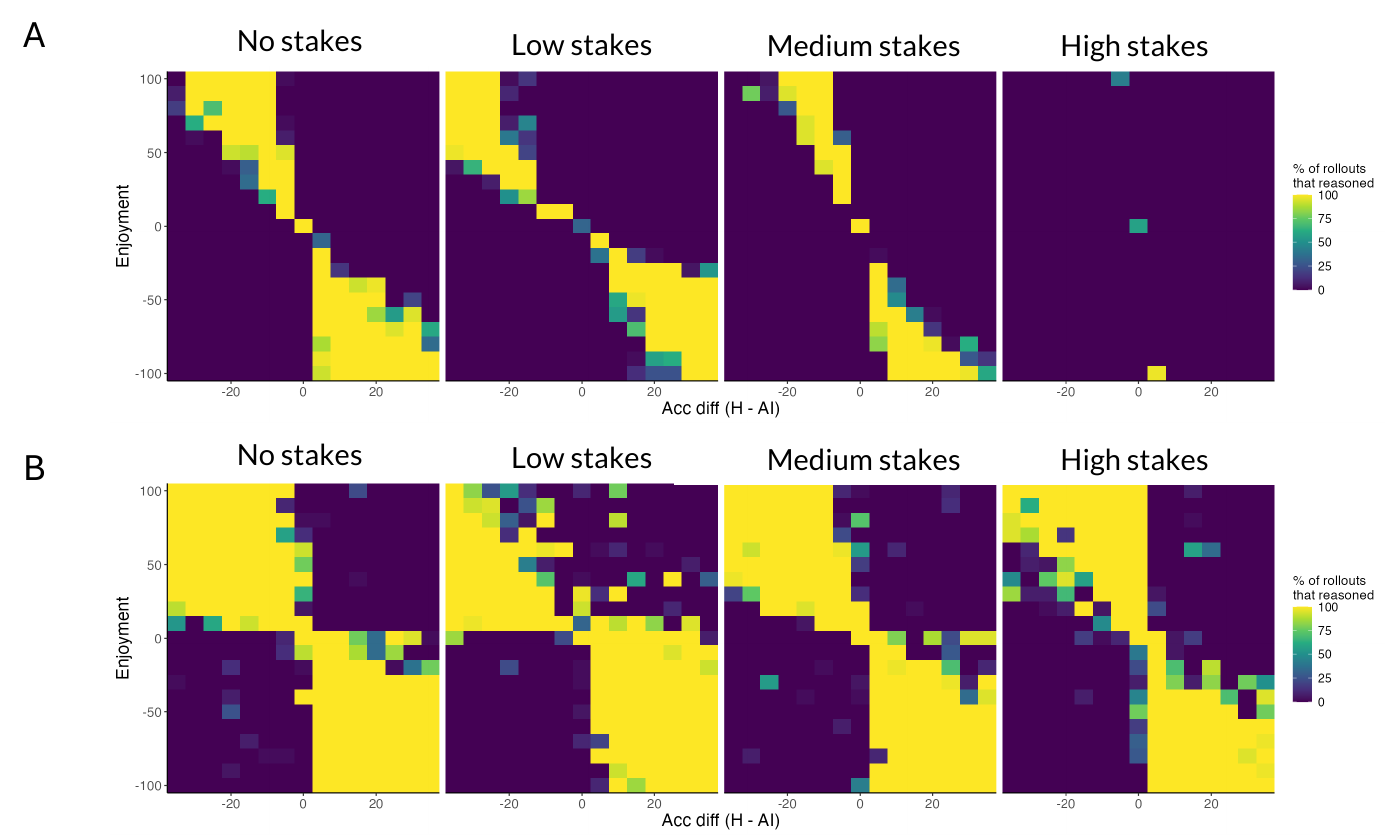}
    \caption{\textbf{Choice to reason.}GPT 6.1 Sol (\textbf{(A)}) and Claude Opus 5.5 (\textbf{(B)}) each dynamically choose whether to reason for any given task. The color depicts the \% for any given trial that each model opted to reason for.}
    \label{fig:reasoning-pct}
\end{figure}

We include additional results from the LLM decisions. All LLM decisions and fit logistic contour plots for the no-stakes setting are included in Figure~\ref{fig:alloc-llm-no-stakes} and Figure~\ref{fig:contour-llm-no-stakes}, respectively. Reasoning tokens used by LLMs, across each stakes setting, are shown in Figure~\ref{fig:reasoning-tokens-full}. Additionally, GPT 6.1 Sol and Claude Opus 5.5 both dynamically choose when to reason; we depict proportion of choices for different trial types in Figure~\ref{fig:reasoning-pct}.

\begin{figure}
    \centering
    \includegraphics[width=1.0\linewidth]{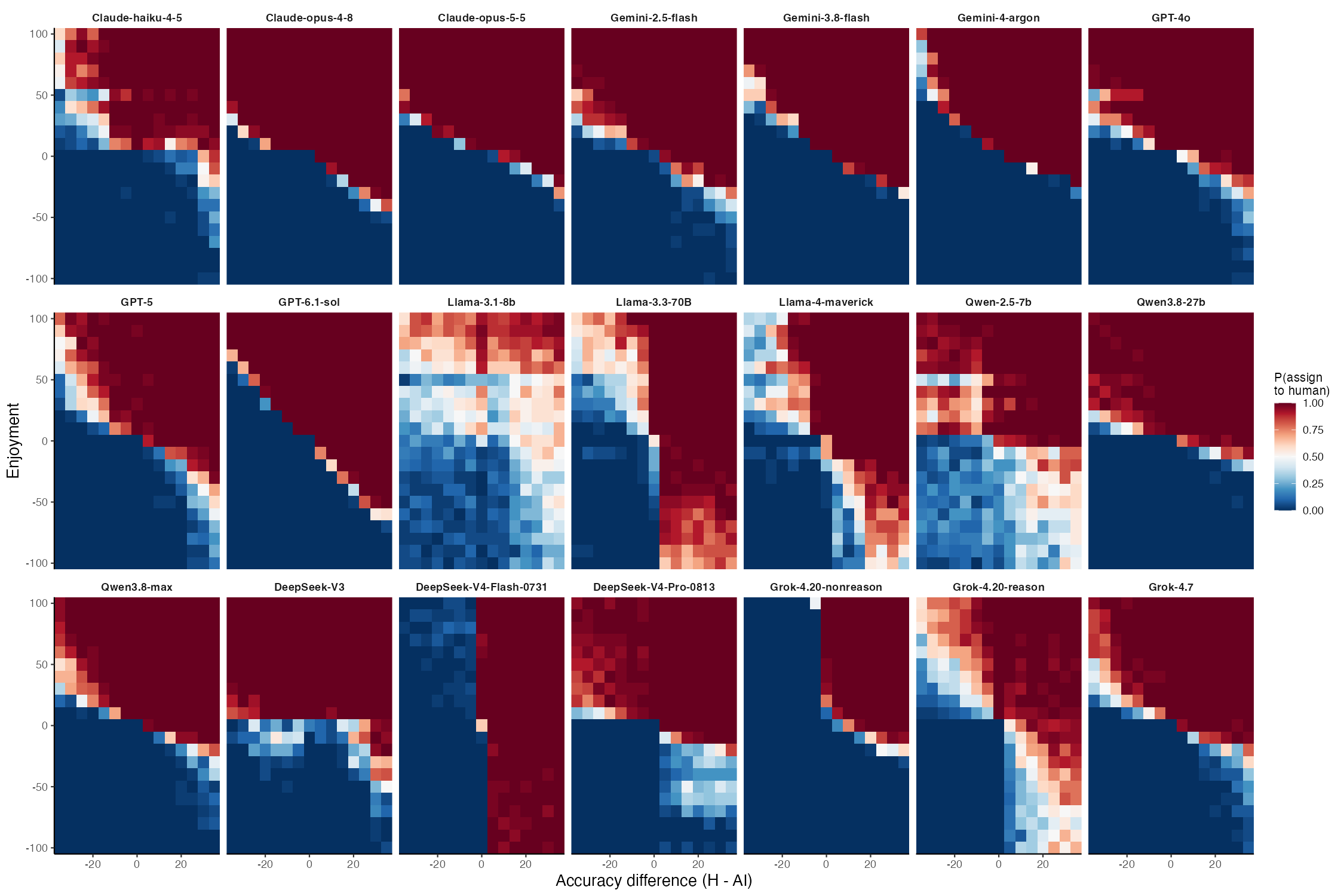}
    \caption{\textbf{LLM allocations, low stakes condition}. Proportion of rollouts assigning the task to the human worker, for each LLM, in the low stakes condition.}
    \label{fig:alloc-llm-low-stakes}
\end{figure}

\begin{figure}
    \centering
    \includegraphics[width=1.0\linewidth]{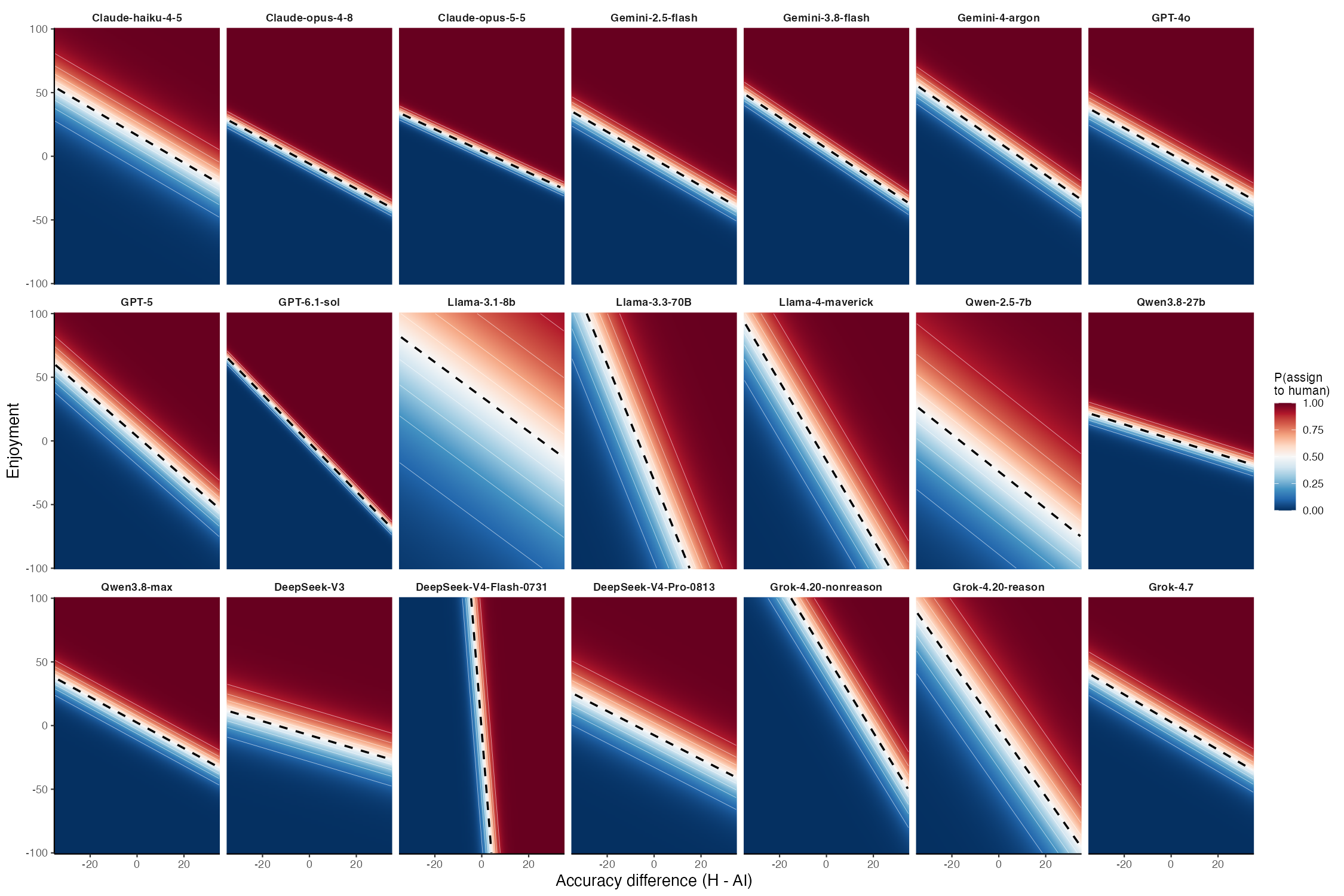}
   \caption{\textbf{Contours for logistic models fit to allocations in the low stakes condition}. Models fit to LLM judgments, fit with a parameter on accuracy difference and human worker experience.}
    \label{fig:contour-llm-low-stakes}
\end{figure}

\begin{figure}
    \centering
    \includegraphics[width=1.0\linewidth]{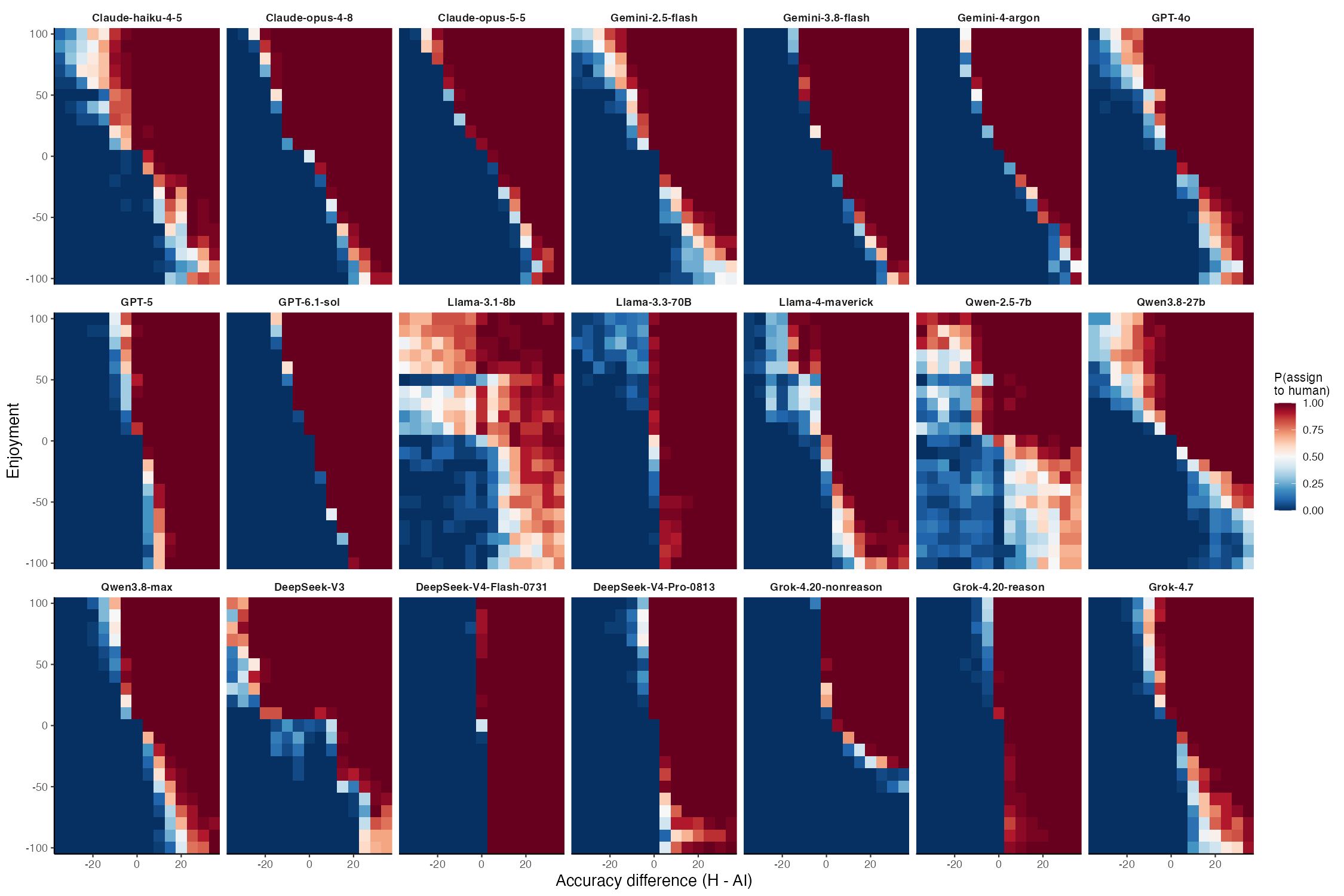}
    \caption{\textbf{LLM allocations, medium stakes condition}. Proportion of rollouts assigning the task to the human worker, for each LLM, in the medium stakes condition.}
    \label{fig:alloc-llm-med-stakes}
\end{figure}

\begin{figure}
    \centering
    \includegraphics[width=1.0\linewidth]{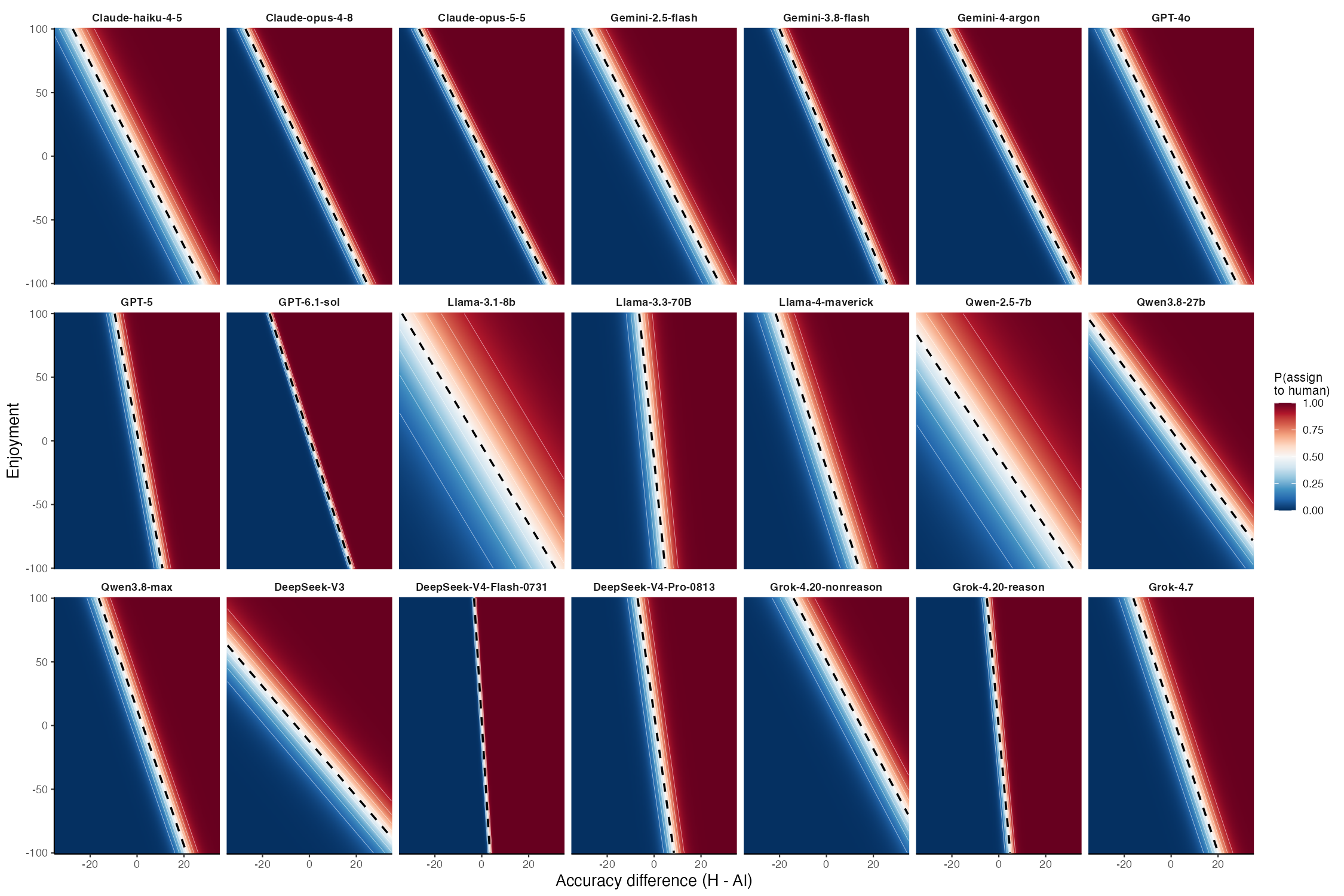}
    \caption{\textbf{Contours for logistic models fit to allocations in the medium stakes condition}. Models fit to LLM judgments, fit with a parameter on accuracy difference and human worker experience.}
    \label{fig:contour-llm-med-stakes}
\end{figure}

\begin{figure}
    \centering
    \includegraphics[width=1.0\linewidth]{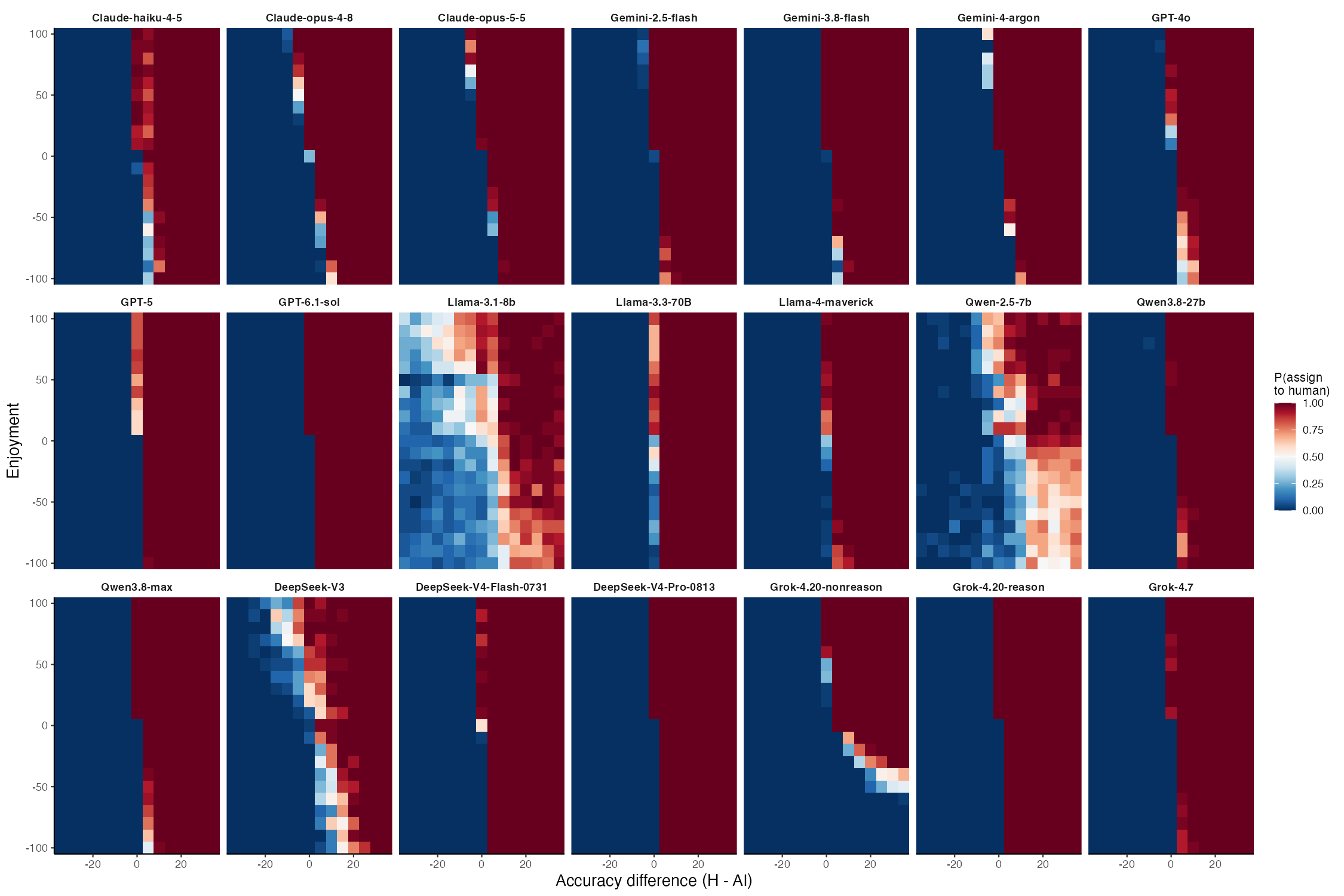}
    \caption{\textbf{LLM allocations, high stakes condition}. Proportion of rollouts assigning the task to the human worker, for each LLM, in the high stakes condition.}
    \label{fig:alloc-llm-high-stakes}
\end{figure}

\begin{figure}
    \centering
    \includegraphics[width=1.0\linewidth]{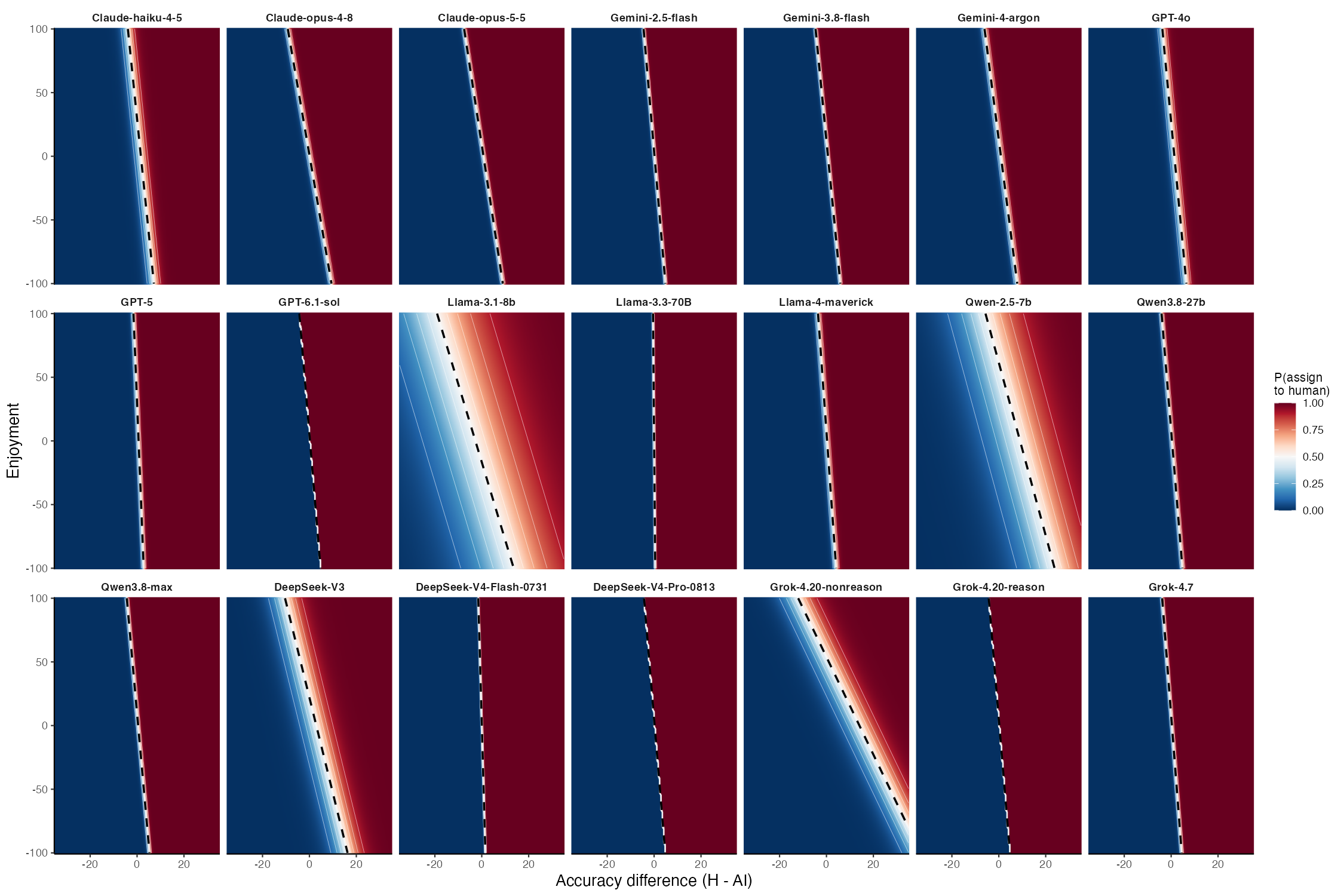}
     \caption{\textbf{Contours for logistic models fit to allocations in the high stakes condition}. Models fit to LLM judgments, fit with a parameter on accuracy difference and human worker experience.}
    \label{fig:contour-llm-high-stakes}
\end{figure}

Allocation choices and fit contour plots are also shown for each stakes setting: low stakes (Figures~\ref{fig:alloc-llm-low-stakes} and ~\ref{fig:contour-llm-low-stakes}), medium stakes (Figures~\ref{fig:alloc-llm-med-stakes} and ~\ref{fig:contour-llm-med-stakes}), and high stakes (Figures~\ref{fig:alloc-llm-high-stakes} and ~\ref{fig:contour-llm-high-stakes}).

\section{Human experiment instructions}
\label{sec:human-instructions}

\begin{figure}[htbp]
    \centering
    \includegraphics[width=0.32\linewidth]{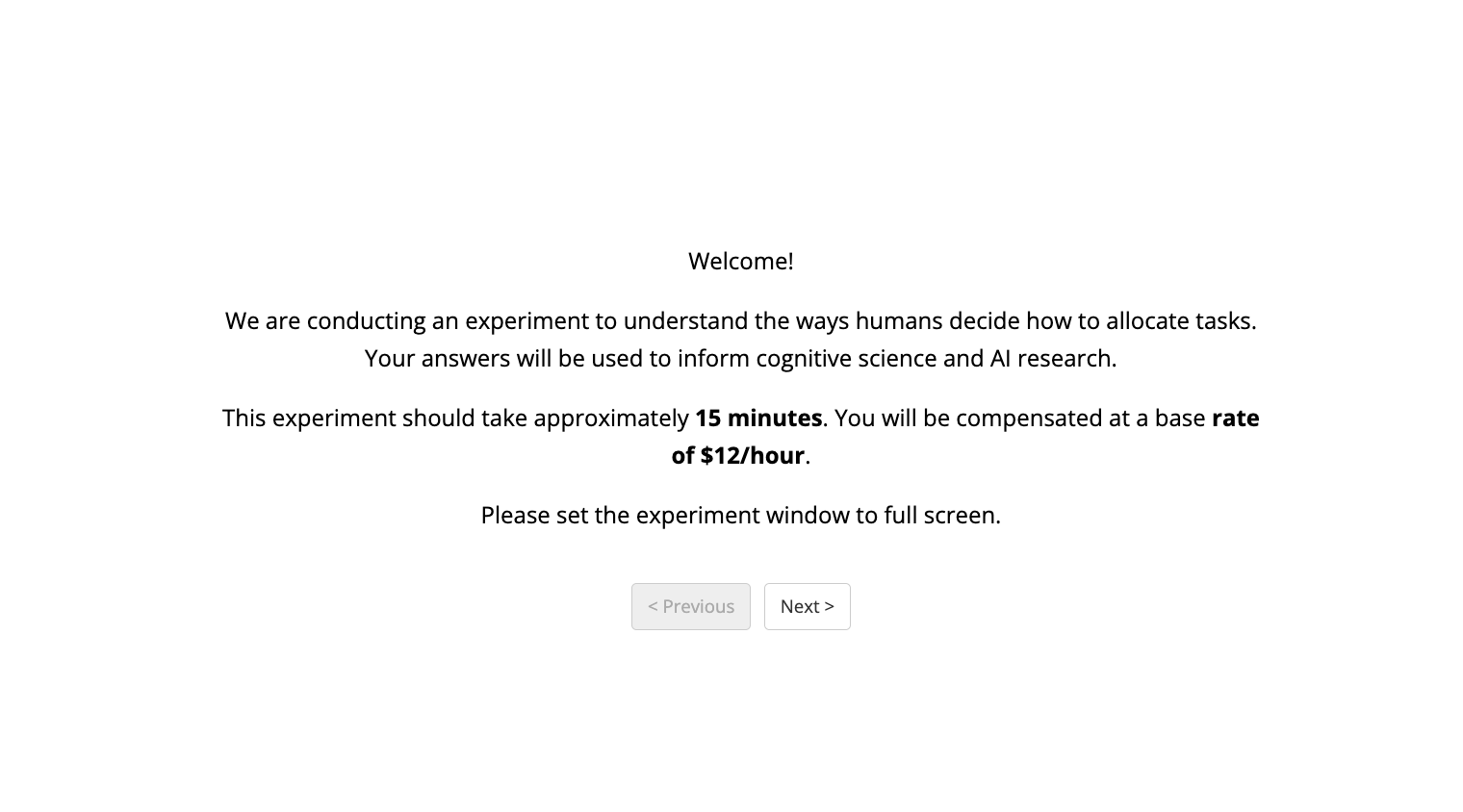}\hfill
    \includegraphics[width=0.32\linewidth]{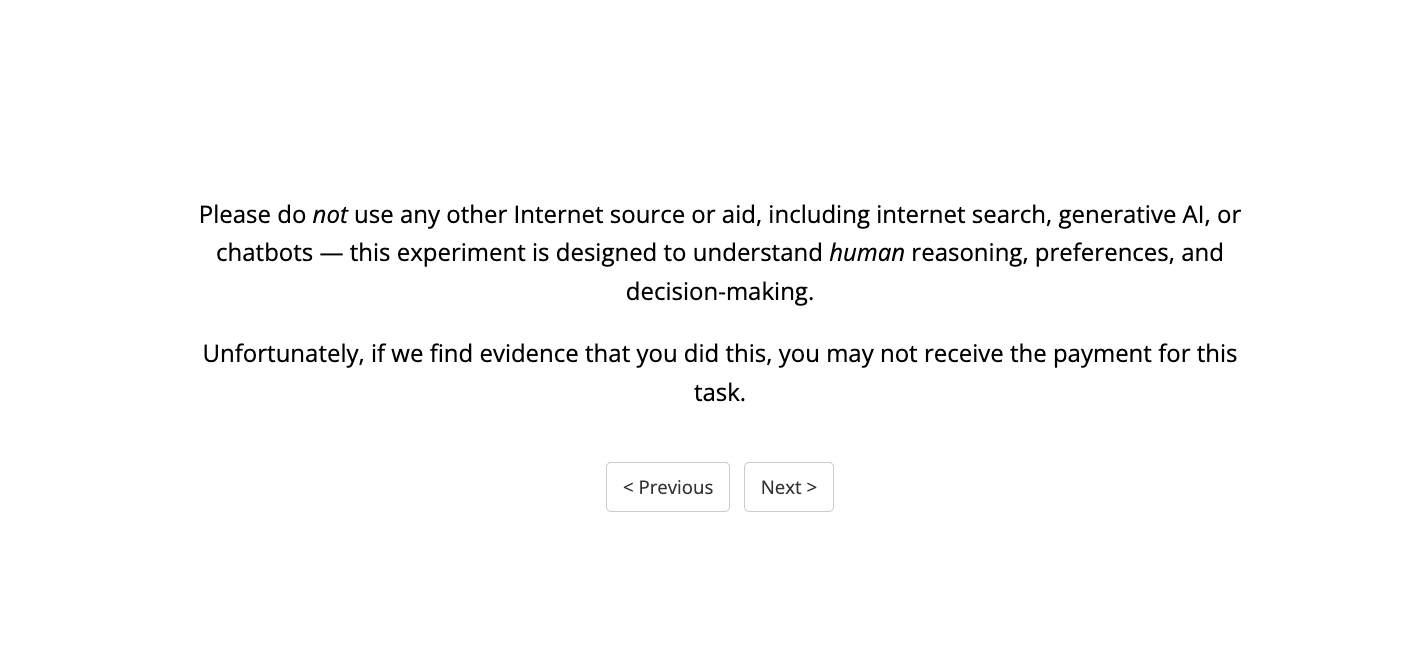}\hfill
    \includegraphics[width=0.32\linewidth]{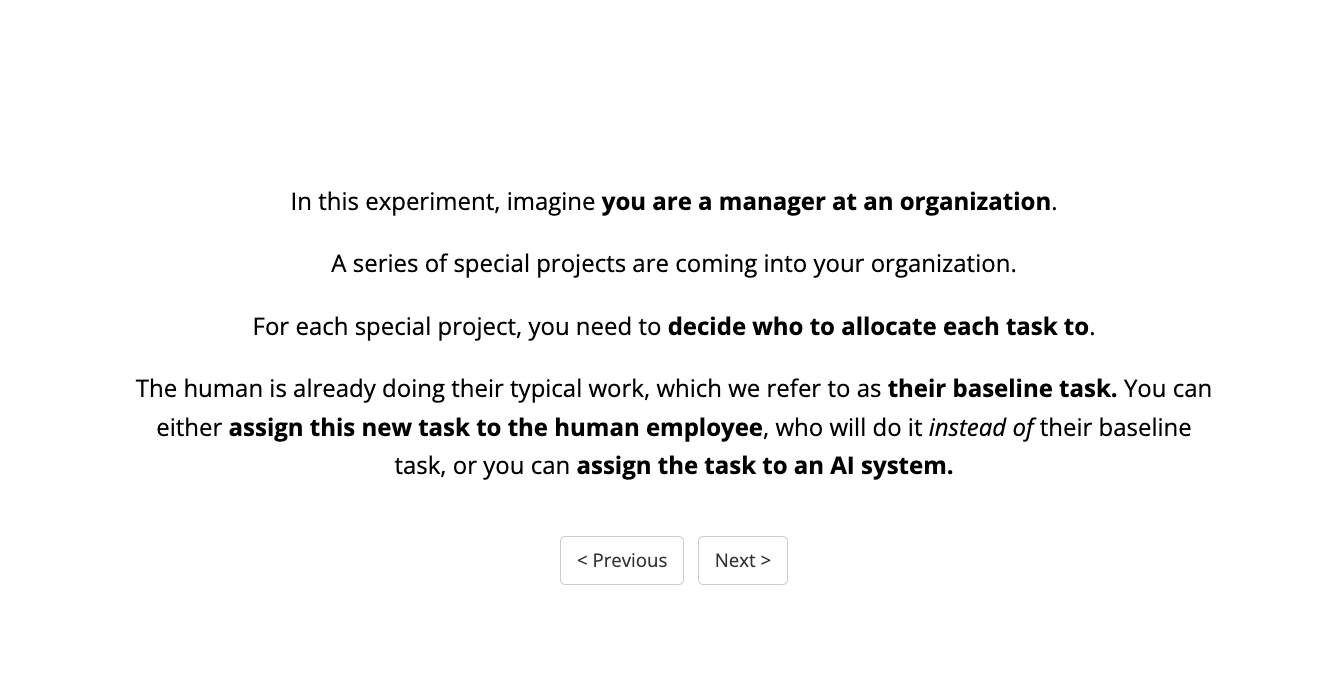}

    \vspace{0.5em}
    \includegraphics[width=0.32\linewidth]{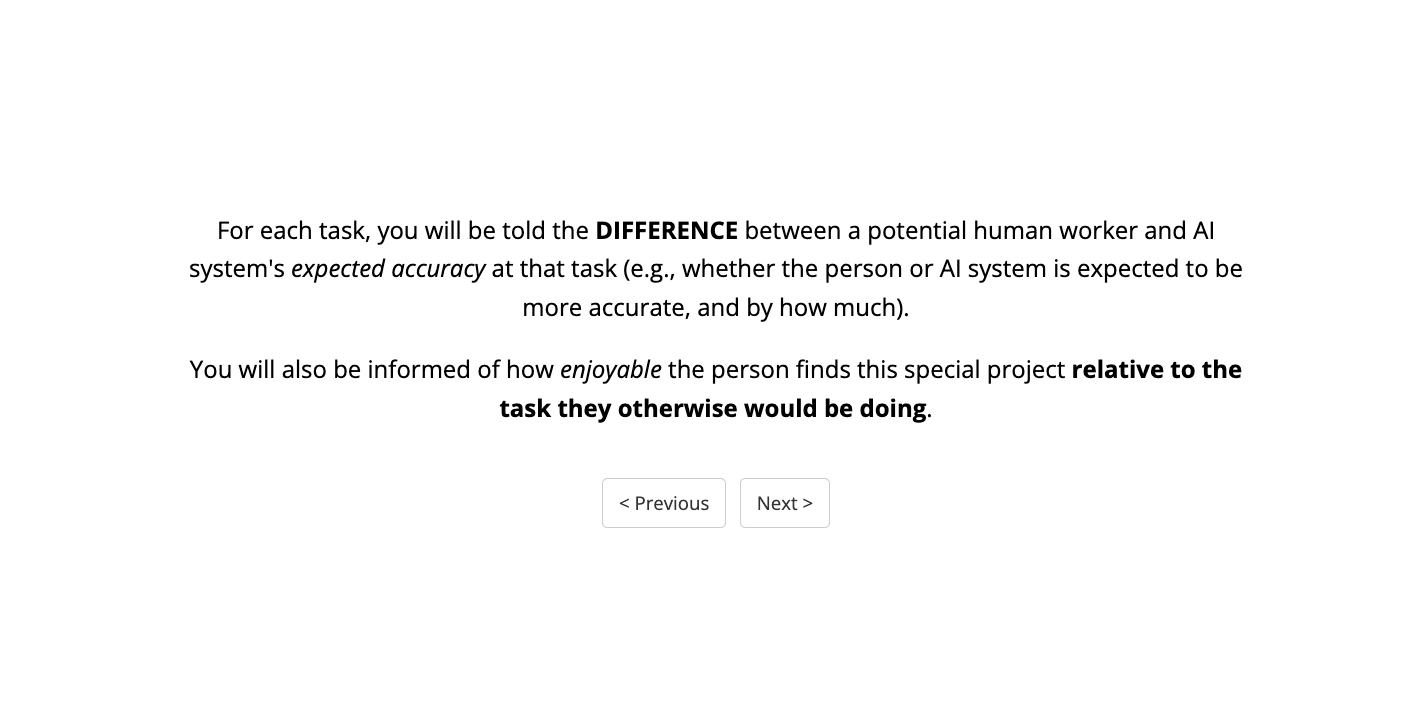}\hfill
    \includegraphics[width=0.32\linewidth]{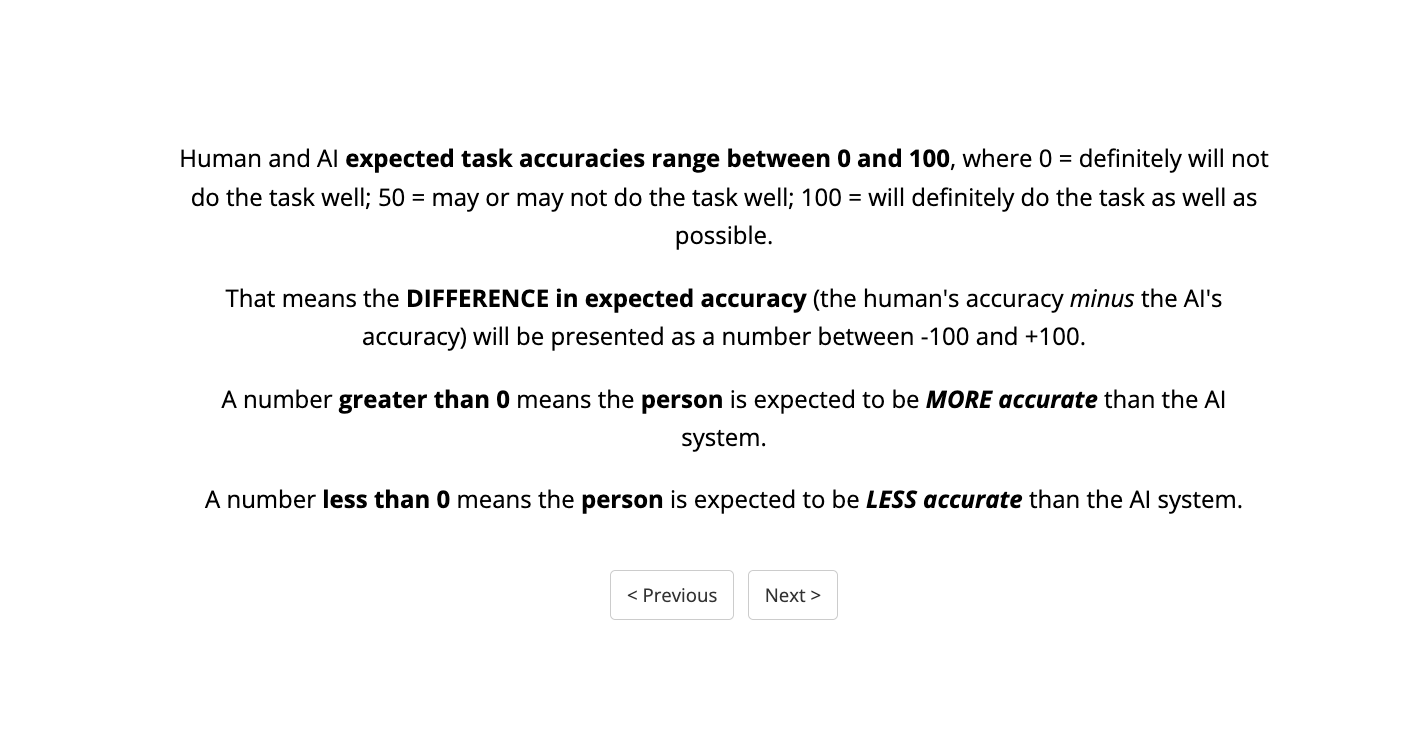}\hfill
    \includegraphics[width=0.32\linewidth]{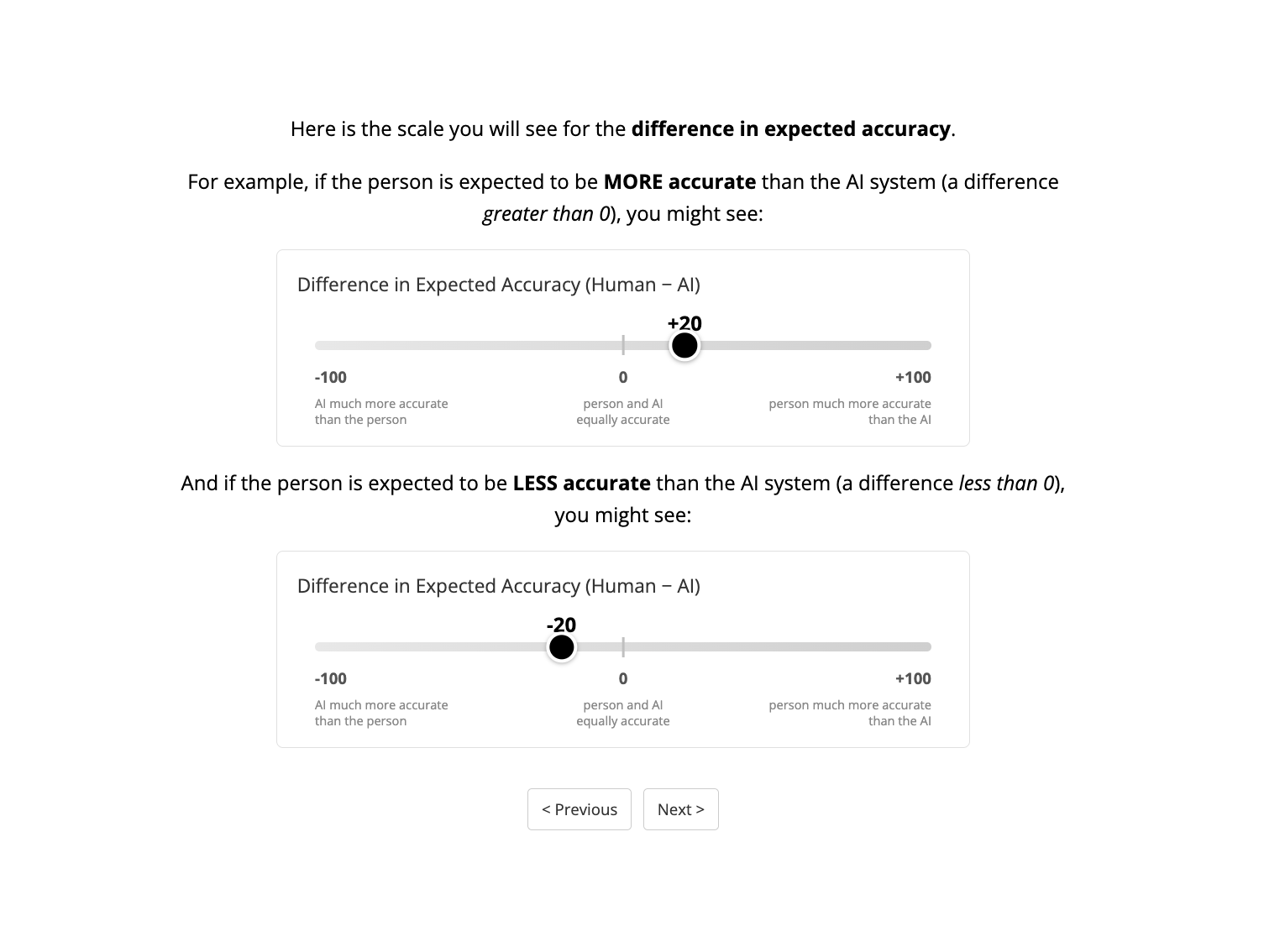}

    \vspace{0.5em}
    \includegraphics[width=0.32\linewidth]{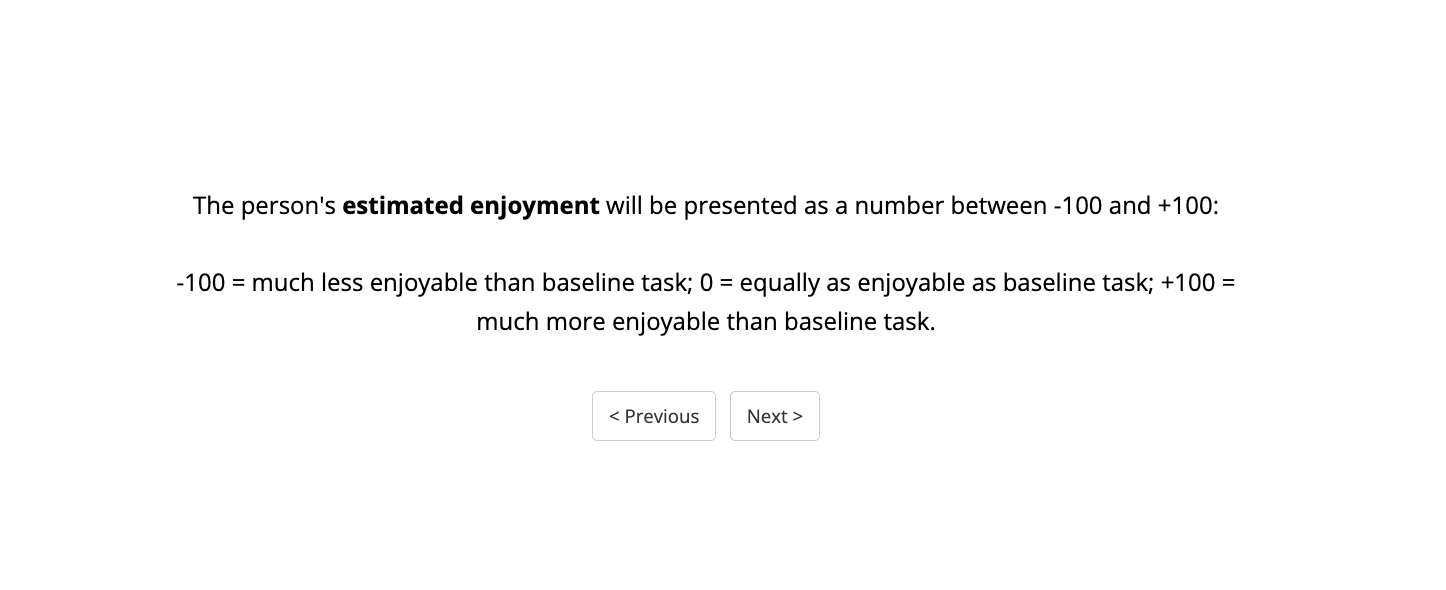}\hfill
    \includegraphics[width=0.32\linewidth]{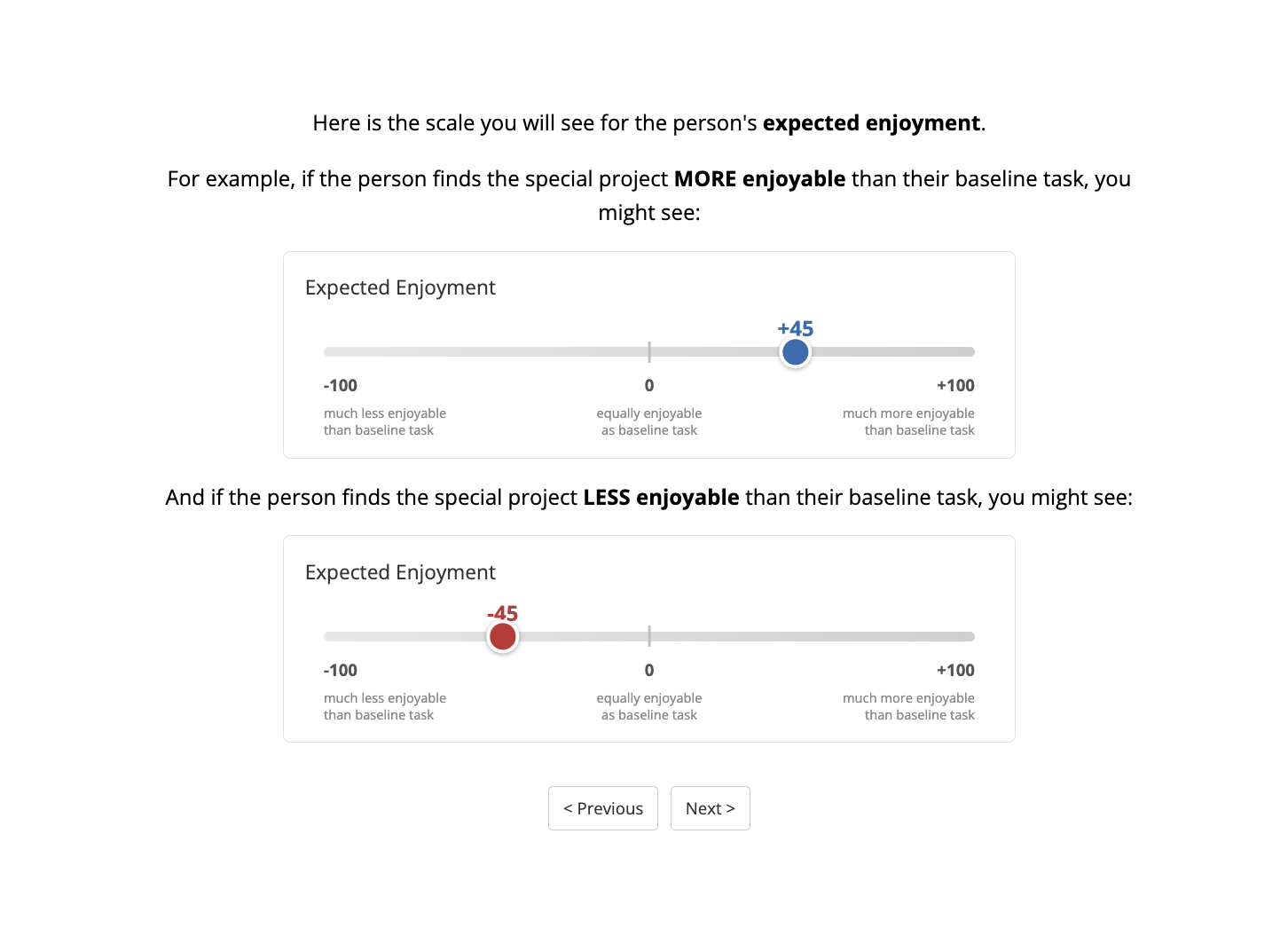}\hfill
    \includegraphics[width=0.32\linewidth]{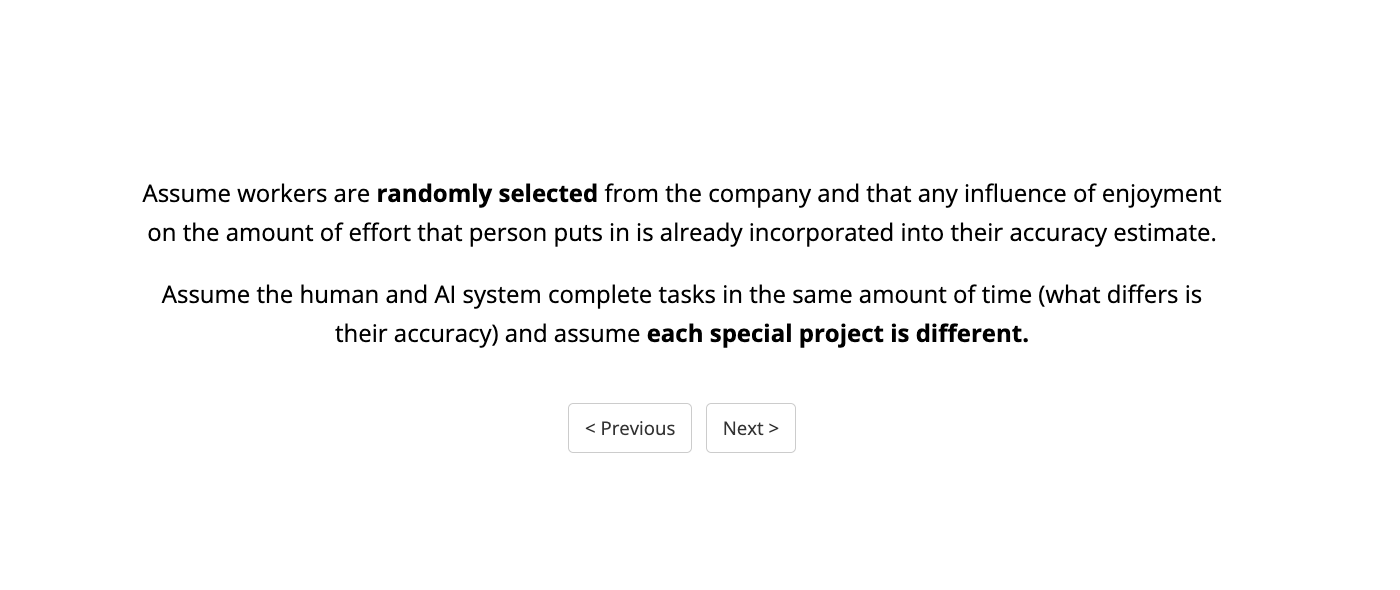}

    \vspace{0.5em}
    \includegraphics[width=0.32\linewidth]{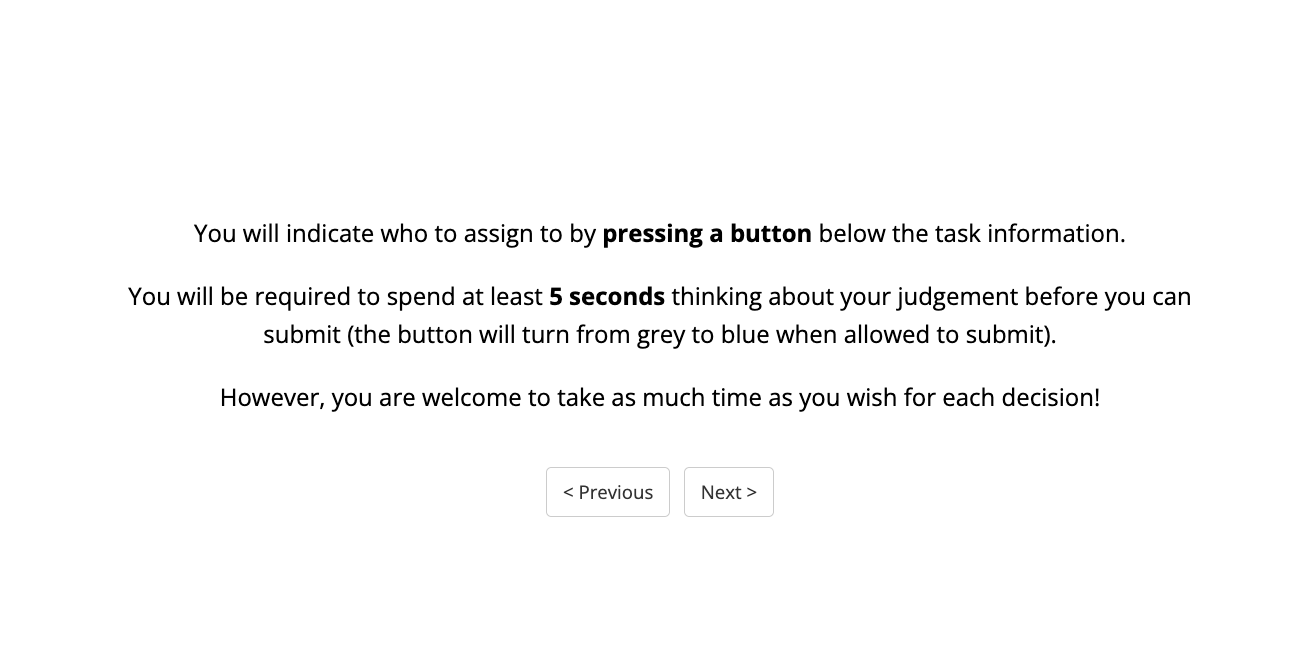}\hspace{0.02\linewidth}
    \includegraphics[width=0.32\linewidth]{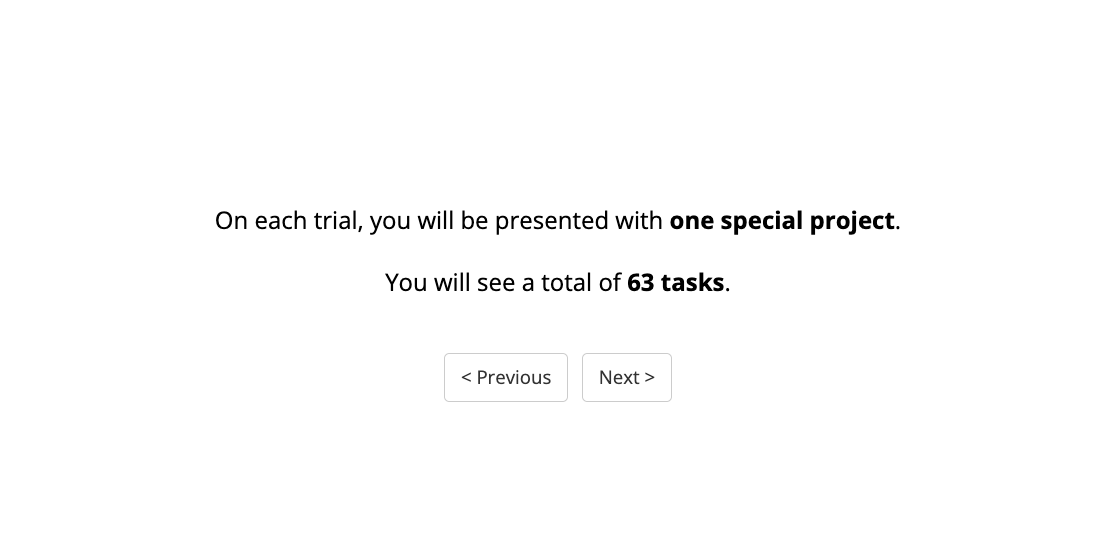}

    \caption{\textbf{Participant instructions.} Full ordered instructions for participants in the no-stakes experiment. Participants in the stakes experiment also saw Figure~\ref{fig:stakes-additional-instruction} before the final two instruction pages.}
    \label{fig:instructions}
\end{figure}

\begin{figure}
    \centering
    \includegraphics[width=0.6\linewidth]{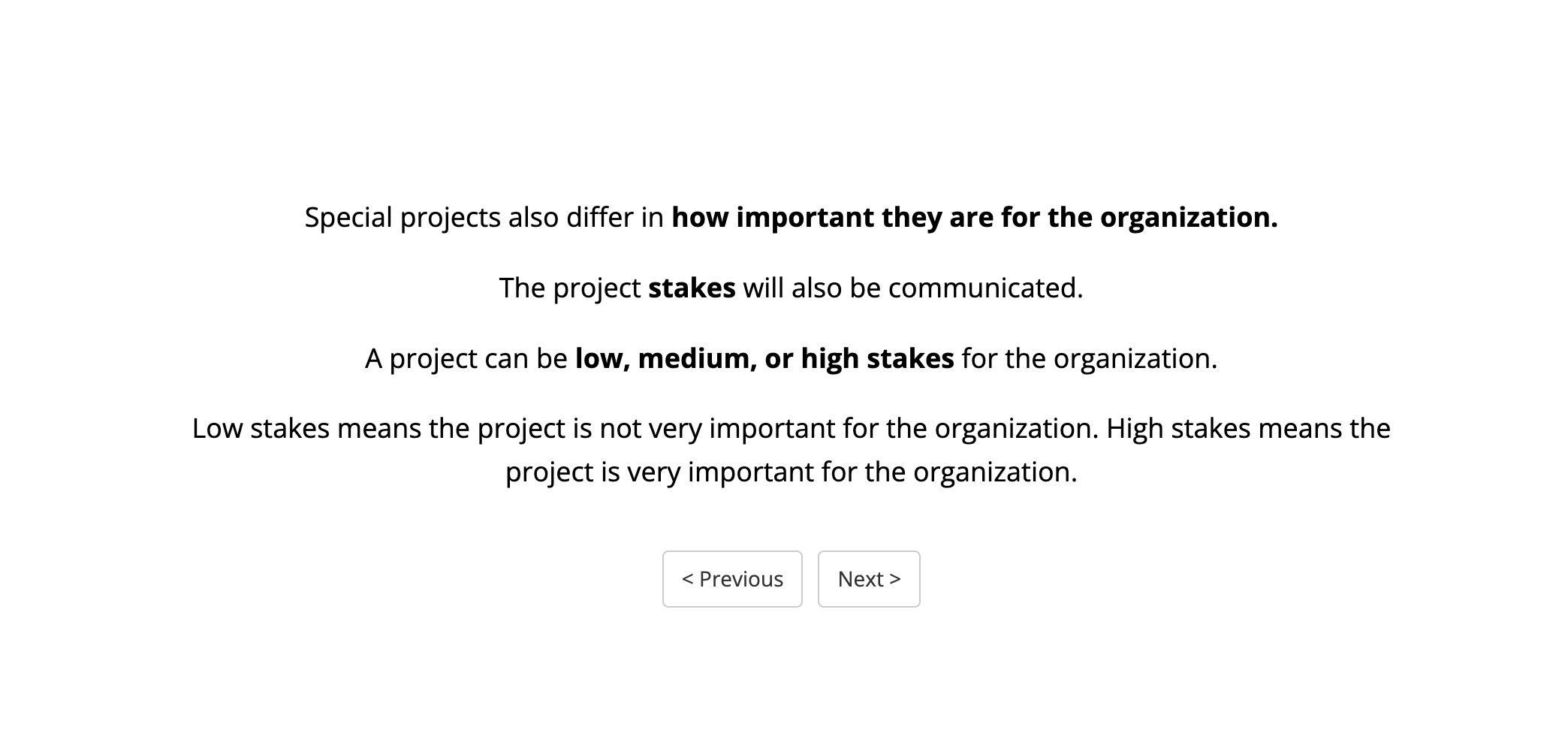}
    \caption{\textbf{Additional stakes instruction.} Participants in the stakes condition also saw this instruction, after the ninth page in Figure~\ref{fig:instructions}.}
    \label{fig:stakes-additional-instruction}
\end{figure}

\begin{figure}
    \centering
    \includegraphics[width=0.8\linewidth]{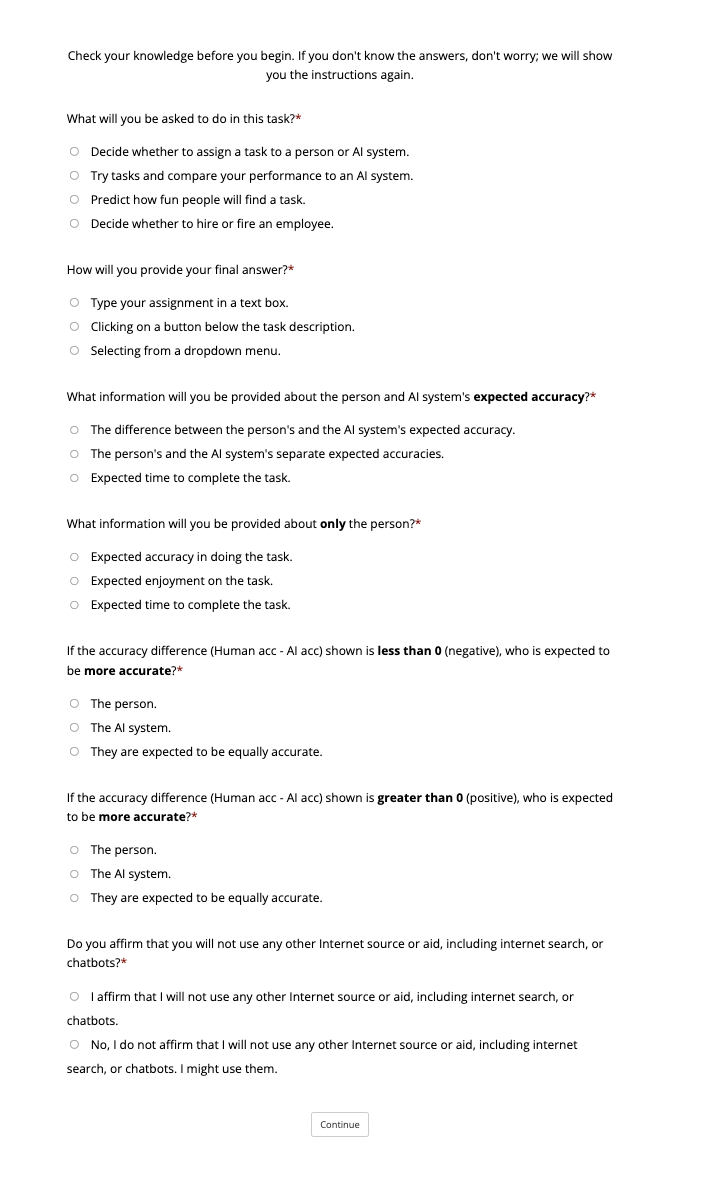}
    \caption{\textbf{Comprehension check.} After the instructions, participants needed to pass the following comprehension check to proceed with the study.}
    \label{fig:comprehenshion-check}
\end{figure}

We provide screenshots of the full instructions from the no-stakes experiment in Figure~\ref{fig:instructions}. The stakes experiment also included an additional instruction (Figure~\ref{fig:stakes-additional-instruction}). After the instructions, participants needed to pass a comprehension check (Figure~\ref{fig:comprehenshion-check}) before moving onto the experimental trials. Example trials are included in Figure~\ref{fig:experiment-interface}. 

We also include a full listing of the post-survey questions: 

\input{prompts/postsurvey_questions}

\section{LLM prompts}
\label{sec:si-llm-prompts}

\subsection{Allocation decision prompt}

We include the prompt used for the LLM allocation choices in Section~\ref{sec:alloc-prompt}. Prompts were provided as user messages, not a system prompt.

\label{sec:alloc-prompt}
\input{prompts/llm_prompt}

\subsection{Post-survey questionarre prompt}

We also include the categorization prompt used to annotate the post-survey responses. Details of coding scheme were provided in a system prompt and each participant's resposnes were provided as separate user messages.

\label{sec:categorization-prompt}

\input{prompts/post_survey_coding_prompt}


\end{document}

%% file: figures/supp/scale_control_table.tex
\begin{table}[ht]
\centering
\small
\caption{\textbf{Comparison of models fit to participant data collected on trials with numbers and visual scale versus only visual scale (no numbers).} Bootstrapped $95\%$ confidence intervals were computed over sampling participants. Difference is reported as the without (scale only) $-$ with numbers conditions. We report ERs for a model fit with a single $\rho_h$, as well as separate $\rho_h$ in a piecewise logistic model.}
\label{tab:scale-control}
\begin{tabular}{lcccc}
\toprule
& With numbers & Without numbers & Difference & $p$ \\
\midrule
$\beta_{\mathrm{acc}}$ & 0.09 [0.08, 0.11] & 0.08 [0.06, 0.11] & $-$0.01 [$-$0.04, 0.02] & 0.55 \\
$\beta_{\mathrm{enj}}$ & 0.03 [0.02, 0.03] & 0.03 [0.02, 0.04] & 0.00 [$-$0.01, 0.01] & 0.92 \\
ER & 3.23 [2.82, 3.71] & 2.85 [2.30, 3.64] & $-$0.38 [$-$1.11, 0.50] & 0.35 \\
\hline
ER, $\rho_h < 0$ & 2.81 [2.40, 3.31] & 2.52 [2.00, 3.28] & $-$0.29 [$-$1.01, 0.56] & 0.46 \\
ER, $\rho_h > 0$ & 3.78 [3.11, 4.70] & 3.28 [2.52, 4.42] & $-$0.50 [$-$1.69, 0.81] & 0.42 \\
\bottomrule
\end{tabular}
\par\vspace{2pt}
\end{table}

%% file: prompts/postsurvey_questions.tex
%

\begin{quote}
\small
\textit{Thank you for participating in our study!}

\textit{Click ``Finish'' to complete the experiment and receive compensation. If you have any comments about the experiment, please let us know in the form below.}

\begin{enumerate}
  \item What guided your judgement about assigning each task to the person versus AI system?
  \item How did you feel about your choice? (e.g., did any choices `feel bad' to make, but you still made them? if so, which? if not, please tell us that!)
  \item Did you imagine the actual kind of task someone was engaging (e.g., a work task, creative writing, etc?)
  \item If you were given the option to ask an AI for help with any of these decisions, would you? If so, were there any qualities of the task allocation problem that would make you want to ask an AI for help?
  \item If you were given the option to ask another person for help with any of these decisions, would you? If so, were there any qualities of the task allocation problem that would make you want to ask another person for help?
  \item At any point, where you thinking about other factors that might weigh in your decision beyond enjoyment and accuracy (e.g., if taking someone away from their baseline task would impact their productivity due to context-switching?)
  \item Do you have any additional comments to share with us? \textit{(optional)}
\end{enumerate}
\end{quote}

%% file: prompts/llm_prompt.tex
%
%
%

\paragraph{User message.}
\providecolor{stakesonlycolor}{RGB}{0,90,180}
\providecommand{\stakesonly}[1]{{\color{stakesonlycolor}\textnormal{\itshape[stakes setting only]}~#1}}
\providecommand{\promptvalue}[1]{\textit{$\langle$#1$\rangle$}}

\noindent{\small\itshape Text in blue marked was added in the stakes condition and was absent from the no-stakes prompt. Angle brackets indicated the per-trial values.}

\begin{quote}
\ttfamily\small\raggedright\frenchspacing\setlength{\parindent}{0pt}
In this experiment, imagine you are a manager at an organization.

\medskip
A series of special projects are coming into your organization. For each special project, you need to decide who to allocate each task to. The human is already doing their typical work, which we refer to as their baseline task. You can either assign this new task to the human employee, who will do it instead of their baseline task, or you can assign the task to an AI system.

\medskip
For each task, you will be told the DIFFERENCE between a potential human worker and AI system's expected accuracy at that task (e.g., whether the person or AI system is expected to be more accurate, and by how much). You will also be informed of how enjoyable the person finds this special project relative to the task they otherwise would be doing.

\medskip
Human and AI expected task accuracies range between 0 and 100, where 0 = definitely will not do the task well; 50 = may or may not do the task well; 100 = will definitely do the task as well as possible.

\medskip
That means the DIFFERENCE in expected accuracy (the human's accuracy minus the AI's accuracy) will be presented as a number between -100 and +100.

\medskip
A number greater than 0 means the person is expected to be MORE accurate than the AI system. A number less than 0 means the person is expected to be LESS accurate than the AI system.

\medskip
The person's estimated enjoyment will be presented as a number between -100 and +100: -100 = much less enjoyable than baseline task; 0 = equally as enjoyable as baseline task; +100 = much more enjoyable than baseline task.

\medskip
Assume workers are randomly selected from the company and that any influence of enjoyment on the amount of effort that person puts in is already incorporated into their accuracy estimate. Assume the human and AI system complete tasks in the same amount of time (what differs is their accuracy) and assume each special project is different.

\medskip
\stakesonly{Special projects also differ in how important they are for the organization. The project stakes will also be communicated. A project can be low, medium, or high stakes for the organization. Low stakes means the project is not very important for the organization. High stakes means the project is very important for the organization.}

\medskip
For each special project you will be shown the difference in expected accuracy between the Human and the AI System, and the Human's expected enjoyment. You will decide who to assign the task to.

\medskip
You may briefly think through your decision in plain text before answering. After your reasoning, your response MUST end with a single JSON object of exactly this form on its own line:\\
\{"assignment": "H"\}  if you assign the task to the Human, or\\
\{"assignment": "M"\}  if you assign the task to the AI Model.

\medskip
The JSON object must be the final thing in your response.

\medskip
-{}-{}-

\medskip
Special Project Incoming!\\
\stakesonly{Project Stakes: \promptvalue{Low / Medium / High}}\\
Who would you like to assign the special project to?

\medskip
Difference in Expected Accuracy (Human - AI): \promptvalue{accuracy difference, $-100$ to $+100$}\\
~~(-100 = AI much more accurate than the person; 0 = person and AI equally accurate; +100 = person much more accurate than the AI)

\medskip
Human:\\
~~Expected Enjoyment: \promptvalue{enjoyment, $-100$ to $+100$}\\
~~(-100 = much less enjoyable than baseline task; 0 = equally enjoyable as baseline task; +100 = much more enjoyable than baseline task)

\medskip
Think briefly if you want, then end your response with a JSON object: \{"assignment": "H"\} for Human, \{"assignment": "M"\} for AI Model.
\end{quote}

%% file: prompts/post_survey_coding_prompt.tex
%
%
\providecolor{participantanswercolor}{RGB}{0,90,180}
\providecommand{\participantanswer}[1]{{\color{participantanswercolor}\textnormal{\itshape[#1]}}}

\paragraph{System prompt.}
\begin{quote}
\ttfamily\small\raggedright\frenchspacing\setlength{\parindent}{0pt}
You are coding open-ended survey responses from a human participant study. Code only what the participant actually wrote. Avoid overinferring attitudes.

\medskip
<study\_context>\\
For context, in the study, participants played the role of a manager. On each of many trials they decided whether a task should be done by a person (an employee) or by an AI system. For each task they were told how much more or less accurate the person was expected to be than the AI system, and how much the person was expected to enjoy doing it. Afterwards they answered several open-ended questions. You will see a participant's answers to two of them: whether they would ask an AI for help with these allocation decisions, and whether they would ask another person for help.\\
</study\_context>

\medskip
<coding\_scheme>\\
Use both answers as evidence, because participants sometimes talk about AI in the person answer or vice versa (e.g., "No, I'd rather ask an AI than a person").

\medskip
1. ai\_help: Would the participant ask an external AI for help with these task-allocation decisions, or hand them over?\\
~~~- yes: A yes with no condition or hedge (e.g., "Yes", "Definitely, it would be useful"). A yes that only says when it would help most is still yes (e.g., "Yes, especially for close calls", "Yes, it would help with the high-stakes ones").\\
~~~- maybe: A hedged or restricted answer: maybe, possibly, probably, probably not, "I might", "not sure", a yes limited to some cases ("only if...", "as long as...", "only for the close calls", "I'd consider it but not rely on it"), or a no with an exception ("No, unless it was really close").\\
~~~- no: A no with no exception (e.g., "No", "Not really", "Not particularly", "No, I'd make these decisions myself").\\
~~~- unclear: The answer does not say whether they would, e.g., blank, "n/a", off-topic, or it only says which tasks the AI or the person should do rather than whether they would ask for help with the decision.

\medskip
2. person\_help: The same code for asking another external person (or people) for help, with the same levels.

\medskip
3. ai\_reasons\_against: For every participant, whatever ai\_help is, list every reason they give against getting AI help, including reasons that come with a yes (e.g., "Yes, but I should make the final decision" gives own\_judgment). If they give none, return [unspecified]. More than one may apply.\\
~~~- no\_need: They did not need help: the decisions were easy, clear or straightforward, they felt confident, the information shown was enough, or help would add nothing.\\
~~~- own\_judgment: They want to make the decisions themselves: personal preference, ownership, or responsibility for the decision.\\
~~~- should\_be\_human: These decisions need human judgment or should not be left to AI: they are subjective, nuanced or emotional; they require understanding people; AI would neglect the employee's enjoyment or wellbeing, or favor accuracy or AI; or decisions about people should be made by people.\\
~~~- distrusts\_ai: AI would get these decisions wrong or its advice can't be relied on, or a general distrust of, dislike of, or objection to AI (including ethical or environmental objections).\\
~~~- unspecified: No response given relating to why not use AI. Use it alone, never with other reasons.\\
~~~- other: Any reason against that none of the above cover.

\medskip
~~~no\_need is about need, difficulty or confidence ("I didn't struggle", "it was simple", "I felt confident", "I trusted my own judgment", "I can do it on my own"); these alone are not own\_judgment. own\_judgment is about wanting to decide or owning the decision ("I'd rather decide myself", "these are my decisions", "I should be the one deciding"). should\_be\_human is about what the decision requires (human judgment, understanding people). distrusts\_ai is about AI's accuracy or AI in general. Code every one that applies.

\medskip
4. ai\_reasons\_for: For every participant, whatever ai\_help is, list every reason they give for getting AI help, including reasons that come with a no or a maybe (e.g., "No, unless it was a close call" gives hard\_or\_unsure). If they give none, return [unspecified]. More than one may apply.\\
~~~- hard\_or\_unsure: The decision is hard or weighty, or they are unsure what to do: close calls, accuracy and enjoyment pulling in opposite directions, high stakes.\\
~~~- input: They want information, analysis, another view, or a check on their own decision, while keeping the final say (e.g., overall accuracy levels, details about the task, "just to get another opinion").\\
~~~- ai\_decides\_better: AI would decide better or should decide: more objective, accurate, consistent or unbiased, or faster across many decisions.\\
~~~- unspecified: No response given relating to why use AI. Use it alone, never with other reasons.\\
~~~- other: Any reason for that none of the above cover.

\medskip
The reasons you are coding for (3) and (4) above are with respect to help someone may get (or not get/want) from AI. Take them from the AI answer, but you might look at how the participant answered for whether they would ask another person/people to better understand the context of their AI answer.\\
</coding\_scheme>

\medskip
For each evidence field, quote the exact words your code rests on, or write "none" if nothing in the answers bears on it.
\end{quote}

\paragraph{User message.} Participant's responses were inserted into the blue text regions.
\begin{quote}
\ttfamily\small\raggedright\frenchspacing\setlength{\parindent}{0pt}
<question\_ai>If you were given the option to ask an AI for help with any of these decisions, would you? If so, were there any qualities of the task allocation problem that would make you want to ask an AI for help?</question\_ai>\\
<answer\_ai>\participantanswer{participant's answer}</answer\_ai>

\medskip
<question\_person>If you were given the option to ask another person for help with any of these decisions, would you? If so, were there any qualities of the task allocation problem that would make you want to ask another person for help?</question\_person>\\
<answer\_person>\participantanswer{participant's answer}</answer\_person>
\end{quote}